\documentclass[journal,twocolumn]{IEEEtran}
\usepackage{slashbox,graphicx,times,amsmath,amssymb,cite,subfigure,stfloats,booktabs, url,multirow,afterpage}
\usepackage[lined,ruled,commentsnumbered]{algorithm2e}
\allowdisplaybreaks

\makeatletter

\newcommand{\Rmnum}[1]{\expandafter\@slowromancap\romannumeral #1@}
\makeatother

\begin{document}
\title{Patch Learning}

\author{Dongrui~Wu and Jerry M. Mendel
\thanks{D.~Wu is with the Key Laboratory of Image Processing and Intelligent Control (Huazhong University of Science and Technology), Ministry of Education, China. He is also with the School of Artificial Intelligence and Automation, Huazhong University of Science and Technology, Wuhan, China. Email: drwu@hust.edu.cn.}
\thanks{J.M. Mendel is with the Ming Hsieh Department of Electrical Engineering, University of Southern California, Los Angeles, CA, USA. He is also with the College of Artificial Intelligence, Tianjin Normal University, Tianjin, China. Email: mendel@sipi.usc.edu.}}

\maketitle

\begin{abstract}
There have been different strategies to improve the performance of a machine learning model, e.g., increasing the depth, width, and/or nonlinearity of the model, and using ensemble learning to aggregate multiple base/weak learners in parallel or in series. This paper proposes a novel strategy called patch learning (PL) for this problem. It consists of three steps: 1) train an initial global model using all training data; 2) identify  from the initial global model the patches which contribute the most to the learning error, and train a (local) patch model for each such patch; and, 3) update the global model using training data that do not fall into any patch. To use a PL model, we first determine if the input falls into any patch. If yes, then the corresponding patch model is used to compute the output. Otherwise, the global model is used. We explain in detail how PL can be implemented using fuzzy systems. Five regression problems on 1D/2D/3D curve fitting, nonlinear system identification, and chaotic time-series prediction, verified its effectiveness. To our knowledge, the PL idea has not appeared in the literature before, and it opens up a promising new line of research in machine learning.
\end{abstract}


\IEEEpeerreviewmaketitle

\section{Introduction}

Machine learning has been widely used in our everyday life, e.g., face recognition \cite{Zhao2003}, natural language processing \cite{Collobert2011}, recommender systems \cite{Wang2015b}, affective computing \cite{drwuVRST2010,drwuMTALR2019}, brain-computer interfaces \cite{drwuEA2019,drwuiGS2019}, etc. However, training a high-performance machine learning model is usually a challenging and iterative process, relying on experience and trial-and-error: a simple model is first designed; if its performance is not satisfactory, then some remedies are taken to enhance it.

\begin{figure}[htbp]\centering
\subfigure[]{\label{fig:Parallel}   \includegraphics[width=\linewidth,clip]{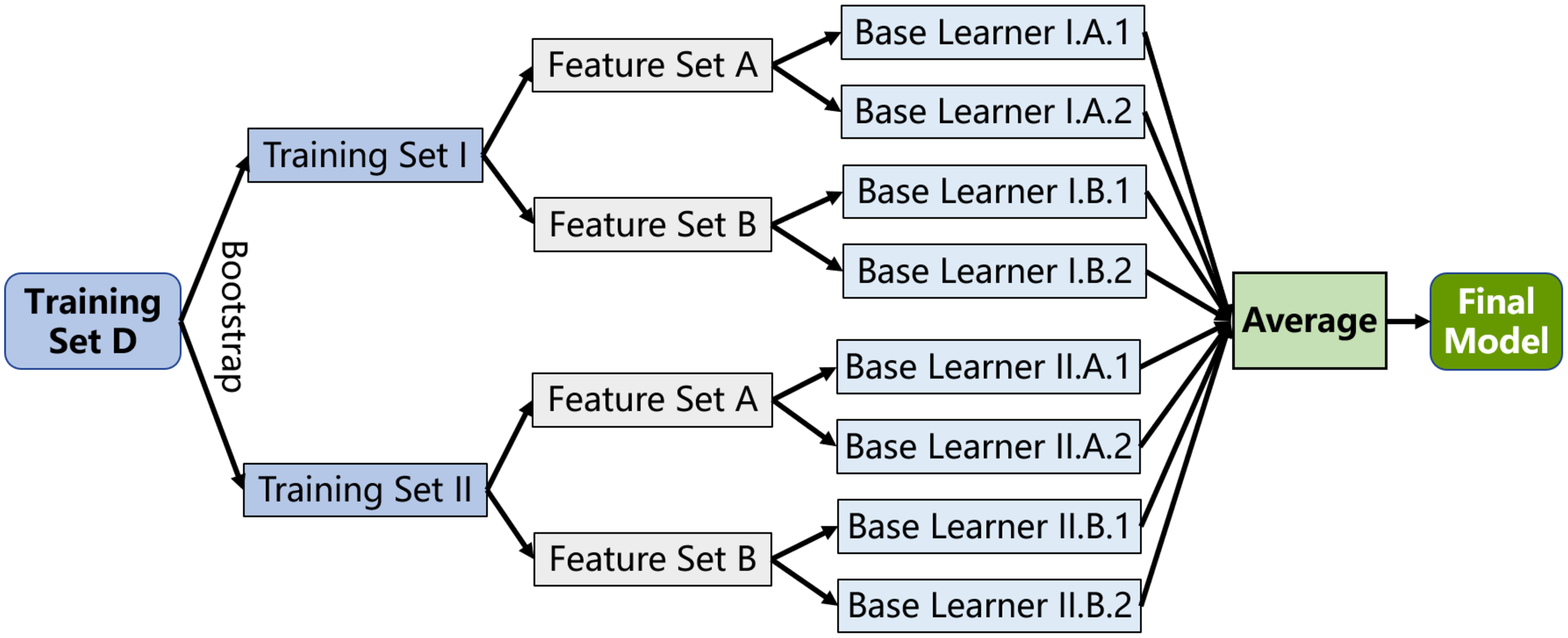}}
\subfigure[]{\label{fig:Serial}    \includegraphics[width=\linewidth,clip]{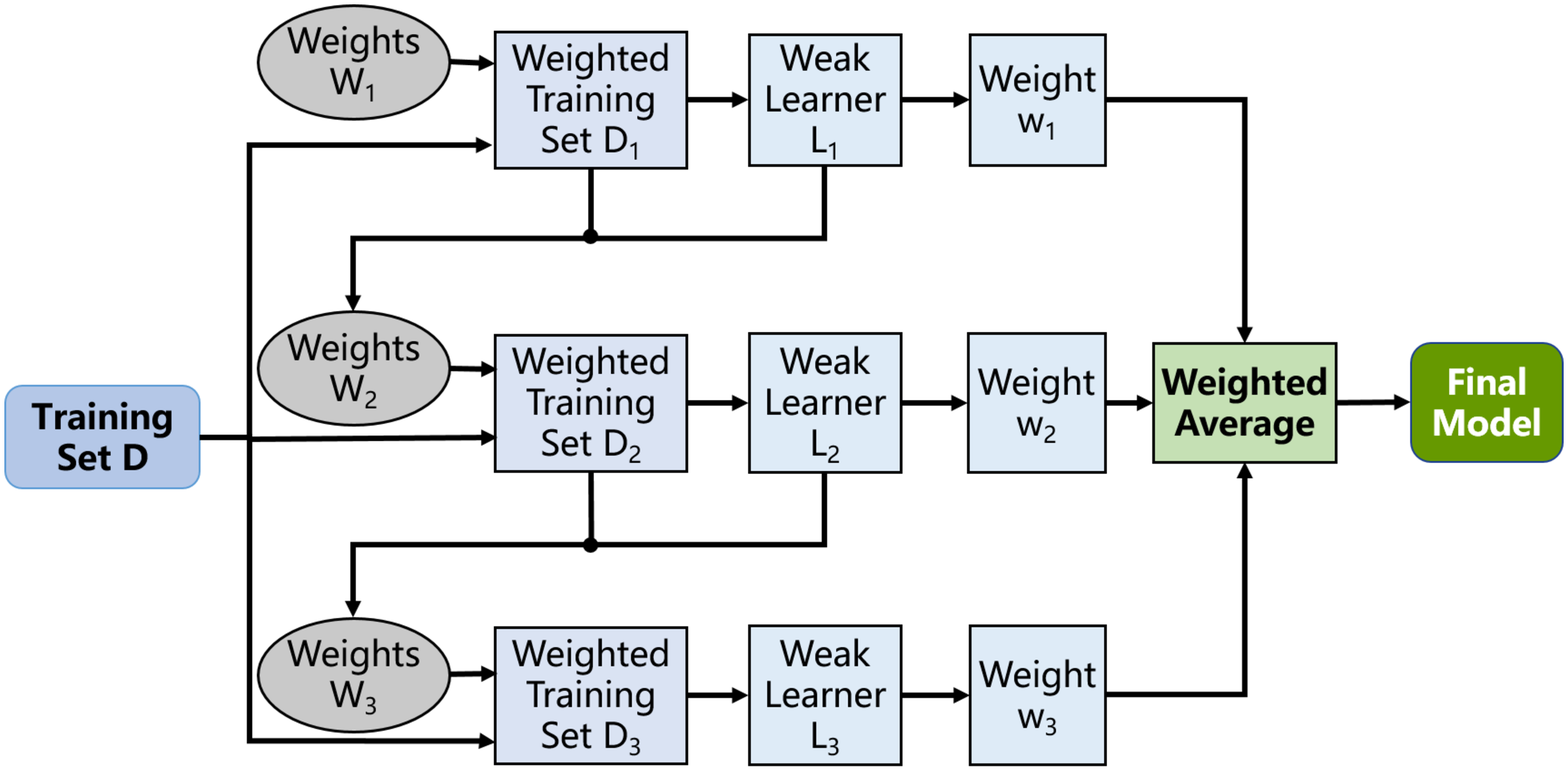}}
\subfigure[]{\label{fig:Patch}    \includegraphics[width=.8\linewidth,clip]{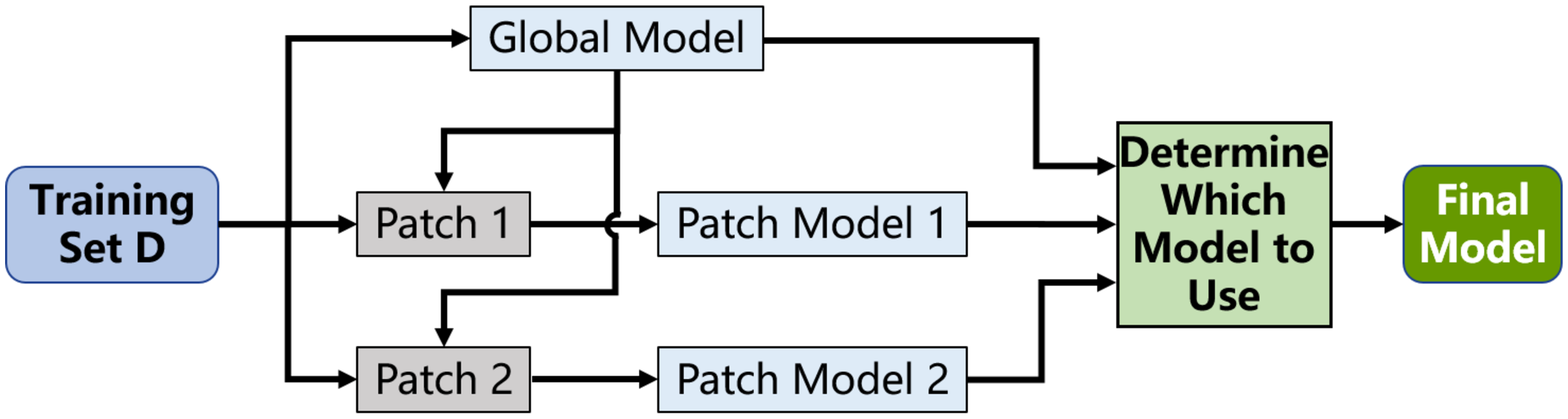}}
\caption{Strategies for connecting multiple simple models for better performance. (a) parallel ensemble learning; (b) serial ensemble learning (AdaBoost \cite{Freund1997a}); and, (c) PL.} \label{fig:EL}
\end{figure}

There have been different strategies to enhance the performance of an under-performing machine learning model:
\begin{enumerate}
\item \emph{Use a single deeper model}. For example, when the performance of a simple multi-layer perceptron neural network \cite{Bishop1995} is not satisfactory, a deep learning model \cite{LeCun2015,He2016}, which has tens or even hundreds of layers, can then be trained. When the performance of a conventional fuzzy system is not enough, a hierarchical fuzzy system \cite{Raju1991} with multiple layers can then be designed.

\item \emph{Use a single broader (wider) model}. For example, when the performance of a simple multi-layer perceptron neural network is not enough, more nodes can be added to each hidden layer \cite{Bishop1995}, or enhancement nodes can be added to convert it into a broad learning system \cite{Chen2018b,drwuMVBLS2019}. When the performance of a simple fuzzy system is not enough, more membership functions (MFs) or rules can be added to make it wider \cite{Mendel2017}; or, in other words, to sculpt the state space more finely \cite{Mendel2018}.

\item \emph{Use a single more non-linear model}. For example, when the performance of a simple linear regression model is not enough, a non-linear regression model like support vector regression using the radial basis function kernel \cite{Smola2004} can be used. When the performance of a simple fuzzy system with constant rule consequents are not enough, a more non-linear Takagi-Sugeno-Kang (TSK) fuzzy system \cite{Takagi1985}, whose rule consequents are linear or non-linear functions of the inputs, can be considered. When the performance of a type-1 fuzzy system is not enough, a more non-linear interval type-2 fuzzy system \cite{drwuEAAI2006,drwuISA2006} can be considered. Note that generally a deeper (broader) model has more non-linearity than a shallower (narrower) model. So, the previous two strategies are also implicitly included in this one.

\item \emph{Connect multiple simple models (usually called base learners) in parallel}. This is a classic idea in ensemble learning \cite{Zhou2012}. For example, when the performance of a base learner is not enough, multiple base learners can be generated in parallel from different partitions of the training data (e.g., bootstrap \cite{Efron1993}, or cross-validation), from different combinations of features (e.g., random forests \cite{Breiman2001}), and/or using different machine learning approaches (e.g., neural networks, fuzzy systems, decision trees, etc.). These base learners can then be aggregated using majority voting (for classification) or averaging (for regression) for better and more robust performance. An illustration of the parallel ensemble learning approach is shown in Fig.~\ref{fig:Parallel}.

\item \emph{Connect multiple simple models (usually called weak learners) in series}. This is another classic idea in ensemble learning. For example, when the performance of a weak learner is not enough, multiple weak learners can be generated in series, each focusing on the hard examples that previous weak learners cannot learn correctly [e.g., AdaBoost \cite{Freund1997a}, illustrated in Fig.~\ref{fig:Serial}], or directly compensating the training error made by previous weak learners (e.g., gradient boosting machine \cite{Friedman2001}).
\end{enumerate}

This paper proposes \emph{patch learning} (PL), which connects multiple simple models both in parallel and in series to improve the learning performance, as illustrated in Fig.~\ref{fig:Patch}. It first trains a global model using all training data, identifies the input regions that give rise to large learning errors, and then designs a patch model for each such region to reduce the overall learning error. The patch models are parallel to and independent of each other, but they are all generated based on the initial global model (and hence in series to the global model). To our knowledge, this idea has not been explored before. We demonstrate the feasibility of PL using fuzzy systems in five regression problems.

The remainder of this paper is organized as follows: Section~\ref{sect:PL} introduces the general idea of PL, and illustrates it by a simple regression problem. Section~\ref{sect:PLFS} describes in detail how PL can be implemented by fuzzy systems. Section~\ref{sect:experiments} presents five experiments on PL using fuzzy systems to demonstrate its feasibility. Section~\ref{sect:limitations} points out some limitations of the current PL approach, and hence opportunities for future research. Finally, Section~\ref{sect:conclusion} draws conclusions.

\section{PL: The General Idea} \label{sect:PL}

This section introduces the general idea of PL, and illustrates it by a simple example.

\begin{figure}[htbp]\centering
\subfigure[]{\label{fig:1Dpatch}   \includegraphics[width=.47\linewidth,clip]{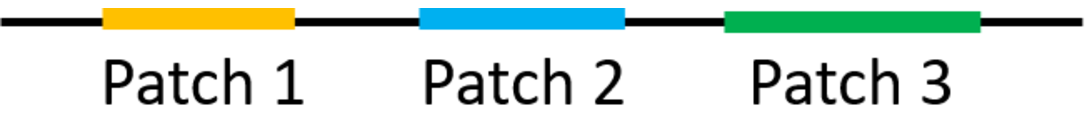}}
\subfigure[]{\label{fig:2Dpatch}    \includegraphics[width=.47\linewidth,clip]{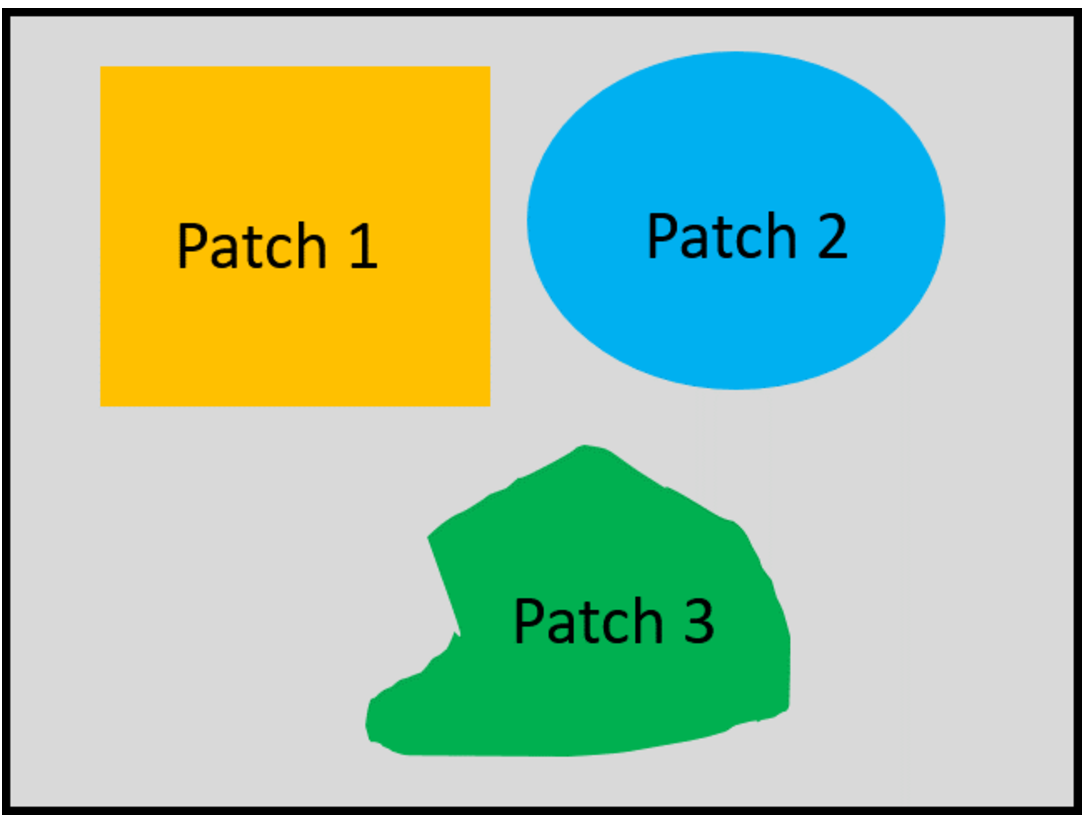}}
\caption{Patches in 1D and 2D input domains.} \label{fig:patchs}
\end{figure}

Formally, we define a \emph{patch} as a connected polyhedron in the input domain. For example, a patch in a 1D input domain is an interval, as shown in Fig.~\ref{fig:1Dpatch}, and a patch in a 2D input domain can be a rectangle, an ellipse, etc., as shown in Fig.~\ref{fig:2Dpatch}. For ease of implementation, in this paper we only consider polyhedra whose number of surfaces (sides) equals the dimensionality of the input domain, and each surface (side) is perpendicular to an axis of the input domain, e.g., an interval in the 1D input domain [Patches~1-3 in Fig.~\ref{fig:1Dpatch}], and a rectangle in the 2D input domain [each side is perpendicular to an axis, such as Patch~1 in Fig.~\ref{fig:2Dpatch}].

\subsection{Steps of PL}

Given $L$, the number of patch models to be trained, PL consists of the following three steps:
\begin{enumerate}
\item Train an initial global model using all training data.
\item Identify $L$ patches from the initial global model, which contribute the most to the learning error, and train a (local) patch model for each such patch.
\item Update the global model using training data that do not fall into any patch.
\end{enumerate}
The rationale for the last step is that, when the default global model is first trained, it is forced to fit all training examples, some of which, e.g., those within the $L$ patches identified in Step (2), may be really difficult to fit. Since these difficult training examples have been handled by the patch models in Step (2), we can update the default global model to fit only the examples outside of the $L$ patches. This should be much easier than fitting all examples together, and hence the root mean squared error (RMSE) of the default global model may be reduced. Since in Step (3) the training examples for the updated global model are all outside of the patches, the updated global model is not suitable for inputs within any of the $L$ patches: although it can output an arbitrary value for such an input, this does not matter, since the patch models are used to handle such cases.

The idea of PL may be intuitively understood by making the following analogy. Consider a sculptor who is sculpting a human figure. After his first pass at this, the sculptor examines the entire figure and notices that improvements need to be made to certain parts of the figure. The sculptor does not throw out the entire figure and begin a new. Instead, he zooms into those parts that need more work, after which he blends in the refined portions of the figure with the rest of the figure. He continues such iterative refinements until he is satisfied with the entire figure. Each patch in PL is analogous to a part in the figure that needs more work. In traditional ensemble learning approaches, usually each base/weak learner also focuses on the entire figure, instead of a part of it.

\subsection{Determine the Optimal Number of Patch Models}

An important question in PL is how to determine $L$, the optimal number of patch models. Increasing $L$ is equivalent to increasing the model complexity in traditional machine learning. So, some analogy also applies here: generally, the training performance increases with $L$, but the test (generalization) performance may first increase and then decrease (indicating overfitting). Inspired from some common practices in traditional machine learning for reducing overfitting \cite{Duda2000,Hastie2009}, there are at least two approaches to determine $L$ in PL:
\begin{enumerate}
\item \emph{Early stopping}. We first partition all available labeled data into a training set and a validation set (these two sets should not overlap). Then, we train different PL models with different $L$ on the training set, and monitor their performances on the validation set. The one with the best validation performance is chosen as the final PL model.
\item \emph{Regularization}. We view $L$ as an indicator of model complexity, and regularize it in the loss function so that $L$ cannot be too large. We then pick the PL model that results in the smallest loss. There could be different choices and implementations of the regularization, similar to the case in traditional machine learning.
\end{enumerate}

The second idea is used in this paper. We mainly consider regression problems, and use the following loss function, which showed satisfactory performance in our experiments:
\begin{align}
\ell= rmse(\mathbb{D})\times (L+1)^\alpha, \label{eq:loss}
\end{align}
where $rmse(\mathbb{D})$ is the training RMSE on the training set $\mathbb{D}$, and $\alpha>0$ is a trade-off parameter between the RMSE and the model complexity. A smaller $\alpha$ prefers a more accurate but maybe more complex model, and a larger $\alpha$ prefers a simpler model with a smaller number of patches, but the RMSE maybe large. Our experiments showed that $\alpha=1/4$ gave satisfactory performance, so $\alpha=1/4$ was used in this paper.

\subsection{PL Illustrated by a Simple Example}

Next, we use a simple regression problem with only one input to illustrate the above procedure.

\begin{figure}[htbp]\centering
\subfigure[]{\label{fig:gPL0}   \includegraphics[width=.47\linewidth,clip]{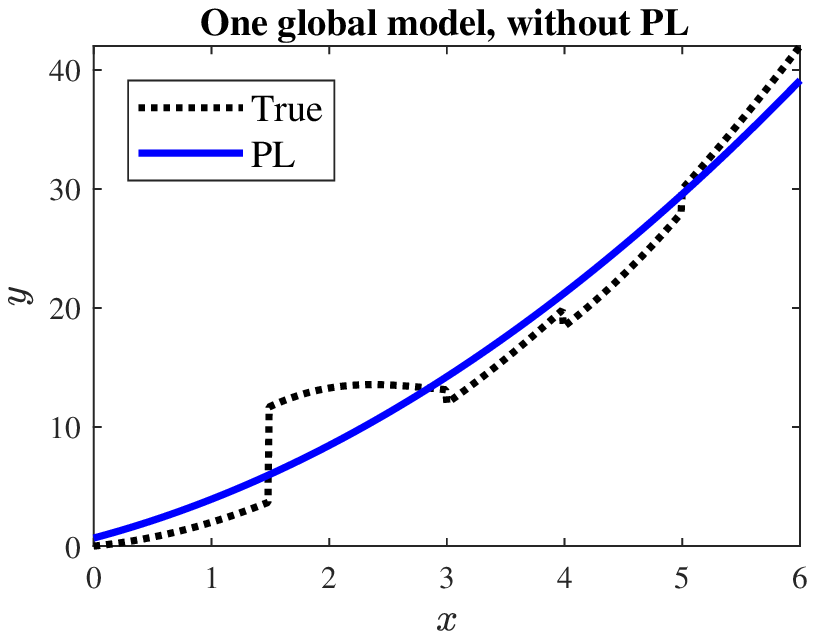}}
\subfigure[]{\label{fig:gRMSE0}    \includegraphics[width=.47\linewidth,clip]{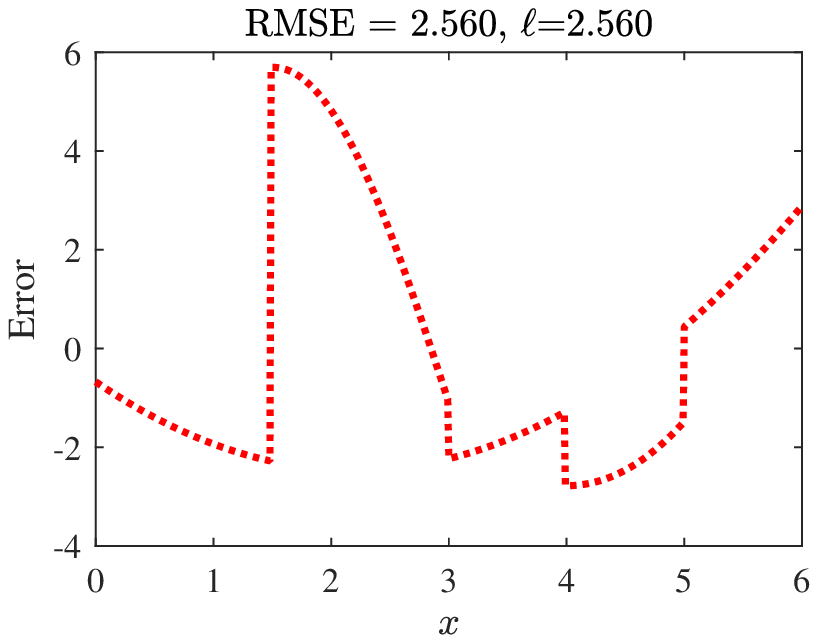}}
\subfigure[]{\label{fig:gPL1}   \includegraphics[width=.47\linewidth,clip]{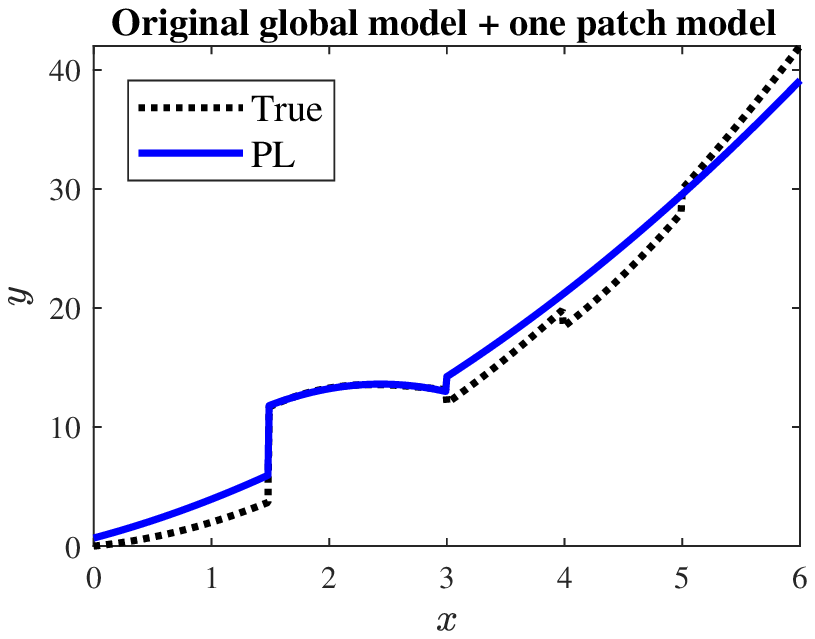}}
\subfigure[]{\label{fig:gRMSE1}    \includegraphics[width=.47\linewidth,clip]{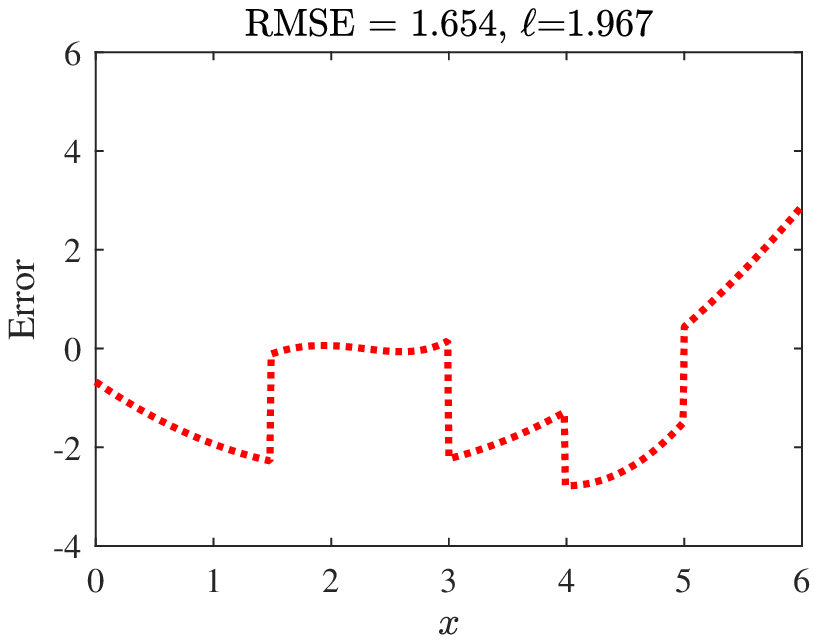}}
\subfigure[]{\label{fig:gPL2}   \includegraphics[width=.47\linewidth,clip]{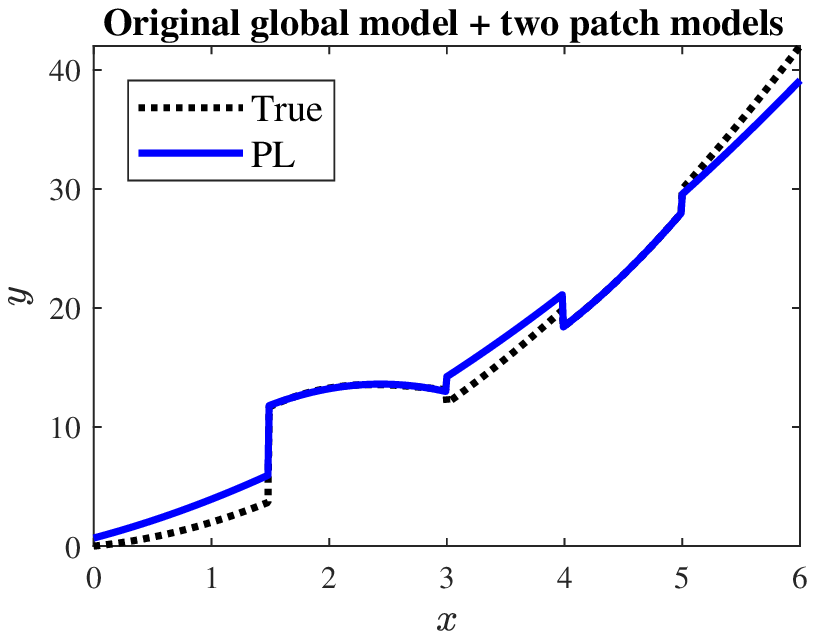}}
\subfigure[]{\label{fig:gRMSE2}    \includegraphics[width=.47\linewidth,clip]{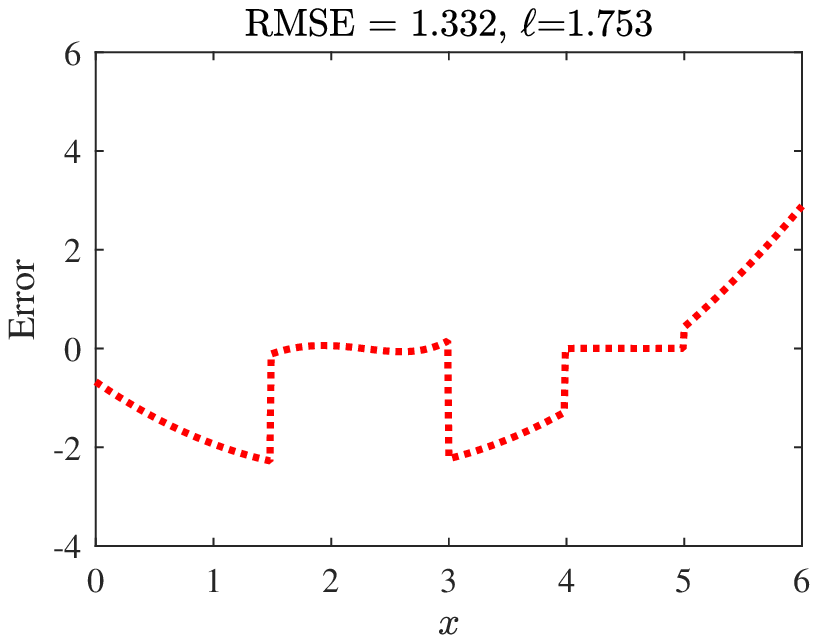}}
\subfigure[]{\label{fig:gPL3}   \includegraphics[width=.47\linewidth,clip]{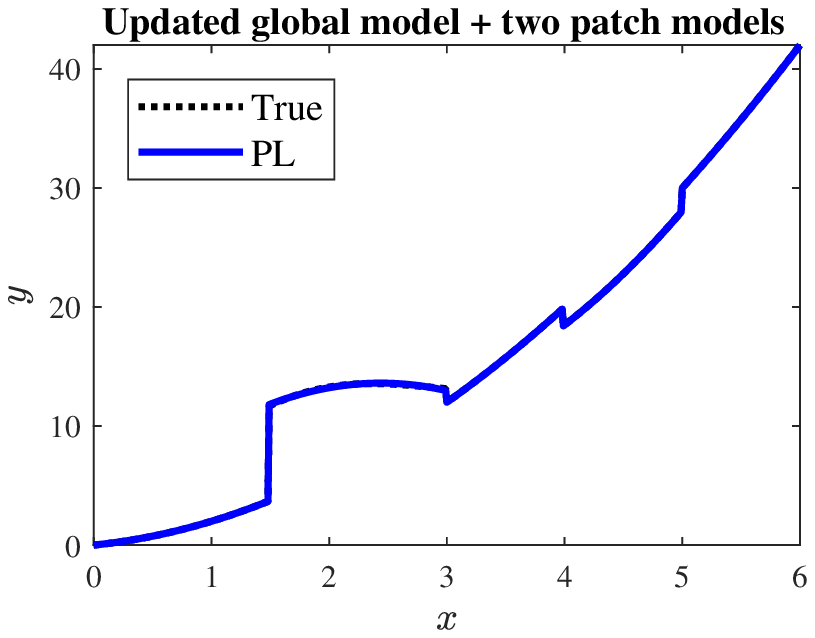}}
\subfigure[]{\label{fig:gRMSE3}    \includegraphics[width=.47\linewidth,clip]{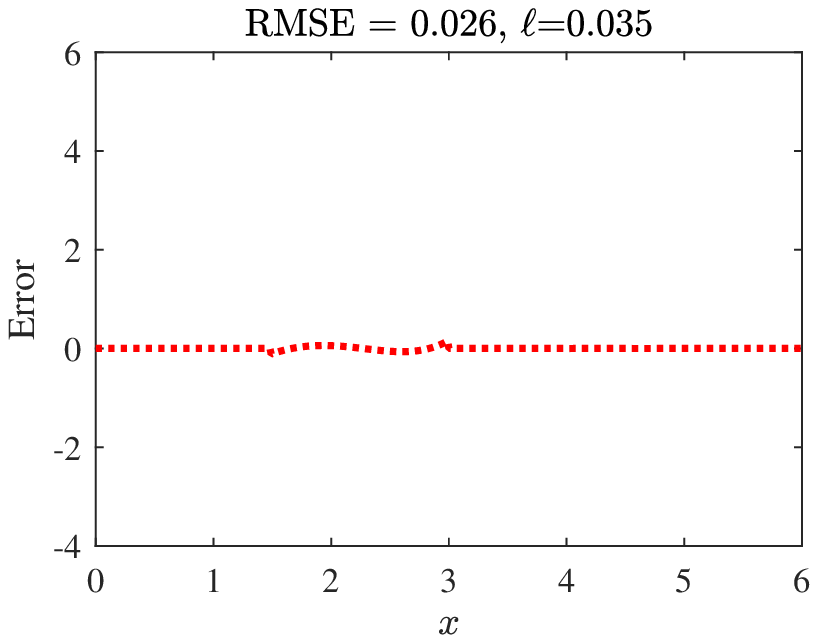}}
\caption{The general idea of PL, illustrated by a simple 1D curve fitting problem. (a) fitting by one global model $f_g(x)$; (b) the fitting error of $f_g(x)$; (c) fitting by the global model $f_g(x)$ plus one patch model $f_1(x)$ in $[1.5,3]$; (d) the fitting error of $f_g(x)$ plus $f_1(x)$; (e) fitting by the global model $f_g(x)$ plus two patch models, $f_1(x)$ in $[1.5,3]$ and $f_2(x)$ in $[4,5]$; (f) the fitting error of $f_g(x)$ plus $f_1(x)$ and $f_2(x)$; (g) fitting by the updated global model $f'_g(x)$ plus the two patch models $f_1(x)$ and $f_2(x)$; (h) the fitting error of the final PL model.} \label{fig:gPL}
\end{figure}

Assume we have $N=601$ training examples $(x_n,y_n)$, $n=1,...,N$, generated from the unknown function
\begin{align}
y=\left\{\begin{array}{ll}
           x+x^2+8\sin(x), & x\in[1.5,3] \\
           x+x^2+2\sin(x), & x\in[4,5] \\
           x+x^2, & \mbox{otherwise}
         \end{array}\right. \label{eq:g}
\end{align}
where $x\in[0,6]$ and 601 uniform samples are used. The true relationship between $y$ and $x$ is plotted as the dotted black curve in Fig.~\ref{fig:gPL0}. We would like to build a PL model to fit it. The basic nonlinear regression model used is $y=f(x)=\beta_0+\beta_1x+\beta_2x^2$.

The first step is to build a global model $f_g(x)$ using all $N$ training examples. The fitted model is $f_g(x)=0.68+2.63x+0.63x^2$, plotted as the solid blue curve in Fig.~\ref{fig:gPL0}. The fitting errors are shown in Fig.~\ref{fig:gRMSE0}. Clearly, the RMSE is large, i.e., a single global model cannot fit the data well.

By visual examination of Fig.~\ref{fig:gRMSE0}, we can see that the fitting errors are large when $x\in[1.5,3]$. So, the next step is to build a patch model for $x\in[1.5,3]$. Using only the training examples within this patch, we obtain the first patch model $f_1(x)=1.65+9.81x-2.01x^2$. $f_g(x)$ and $f_1(x)$ together reduce the training RMSE from $2.560$ to $1.654$, and the loss $\ell$ from $2.560$ to $1.967$, as shown in Fig.~\ref{fig:gRMSE1}. The fitting accuracy is improved, but the RMSE may still be too large. So, a second patch model may be needed.

By visual examination of Fig.~\ref{fig:gRMSE1}, we can see that the fitting errors are large when $x\in[4,5]$. So, the next step is to build a patch model for $x\in[4,5]$. Using only the training examples within this patch, we obtain the second patch model $f_2(x)=19.29-8.03x+1.96x^2$. $f_g(x)$, $f_1(x)$ and $f_2(x)$ together further reduce the training RMSE to $1.332$, and the loss to $1.753$, as shown in Fig.~\ref{fig:gRMSE2}.

Assume at this point we do not want to add more patches. The final step is then to update the default global model, using all training examples outside of the two patches.  The updated global model is $f'_g(x)=x+x^2$, identical to the last case in (\ref{eq:g}). This step reduces the RMSE to $0.026$, and the loss to $0.035$, as shown in Fig.~\ref{fig:gRMSE3}.

For the above simple 1D curve fitting example, the final PL model can be easily represented as:
\begin{align}
y=\left\{\begin{array}{ll}
           f_1(x)=1.65+9.81x-2.01x^2, & x\in[1.5,3] \\
           f_2(x)=19.29-8.03x+1.96x^2, & x\in[4,5] \\
           f'_g(x)=x+x^2, & \mbox{otherwise}
         \end{array}\right. \label{eq:g2}
\end{align}

\begin{figure}[htbp]         \centering
\includegraphics[width=.6\linewidth,clip]{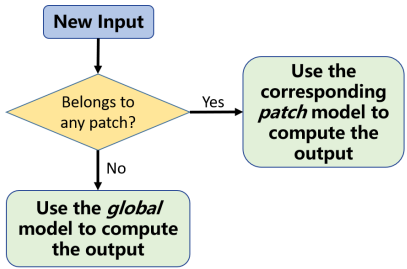}
\caption{The logic for determining which model to use in PL.} \label{fig:PLmodel}
\end{figure}

Generally, once the PL models are trained, the logic for determining which model to use for a new input is shown in Fig.~\ref{fig:PLmodel}. We first determine if the input falls into any patch. If yes, then the corresponding patch model is used to compute the output. Otherwise, the global model is used.

\subsection{Comparison with Boosting}

As pointed out by one of the reviewers, PL can be conceptually understood as a special case of boosting, which uses 0-1 weights instead of continuous weights in training the weak learners: each of the first $L$ weak learners (patch models) assigns weight one to all training examples falling into its corresponding patch, and weight zero to all other training examples; the final weak learner (global model) assigns weight one to training examples that are assigned weight zero by \emph{all} $L$ previous weak learners, and weight zero to training examples that are assigned weight one by \emph{any} of the $L$ previous weak learners. In traditional boosting, for a new input $\mathbf{x}$, the output is a weighted average of the outputs of all $L+1$ weak learners; however, in 0-1 boosting, the final output equals the output of the fired weak learner (only one weak learner is fired for each input).

Note that the idea of patch is critical here, because it determines which weak learner is used in the final prediction. As a simple illustrative example, consider a 0-1 weighted boosting model, which consists of two weak learners but does not use patches: we can train the first weak learner using all training examples, select the 50\% training examples that give the maximum errors, train the second weak learner using them, and finally update the first weak learner using the remaining 50\% training examples. The training can be done without any problem. However, usually the training examples for the two weak learners overlap in the input domain, so we cannot clearly define which weak learner is responsible for which input region. This causes a fatal problem in testing: for a new input $\mathbf{x}$, we cannot determine which weak learner should be used to compute its output, and hence the model cannot be used in practice. On the contrary, in PL each patch model (weak learner) is associated with a patch with explicit boundaries, so we can easily determine which model should be used for a new input.

\section{PL Using Fuzzy Systems} \label{sect:PLFS}

The simple example in the previous section illustrates the general idea of PL. However, generally identifying the patch locations is a very challenging task, and not every problem can be easily visualized. This section introduces how PL can be performed using rule-based fuzzy systems (fuzzy systems, for short), because it is easy to initialize patch candidates in a fuzzy system.

\subsection{The Overall Procedure}

The traditional design of a fuzzy system is global, in the sense that representative training data are used to
optimize the input MFs and consequent parameters over $D_1\times\cdots\times D_M$, where $D_m$ denotes the $m$th input domain ($m=1,...,M$). During the design stage, performance metrics are optimized using all training data.

A PL fuzzy system begins with a globally designed fuzzy system, but then locates the patches in $D_1\times\cdots\times D_M$ which have contributed the most to the performance metrics (e.g., the largest RMSE, the most misclassifications, etc.). A patch fuzzy system is then designed for each such patch using a subset of training data that are in that patch, in such a way that performance metrics are improved within each patch. Finally, the global fuzzy system is updated, using only the remaining training data that have not been used by any patch. The performance metrics will always be better for the PL fuzzy system than for the initial global fuzzy system.

Once this three-stage design is completed, the PL fuzzy system is comprised of the updated global fuzzy system and a collection of patch fuzzy systems. Of course, for each patch fuzzy system, we also need to record its corresponding patch locations.

The procedure for computing the output for a new input has been shown in Fig.~\ref{fig:PLmodel}. More specifically, for each measured input $\mathbf{x}$,
\begin{itemize}
\item If $\mathbf{x}$ belongs to a certain patch, then compute the output $y(\mathbf{x})$ using the corresponding patch fuzzy system.
\item If $\mathbf{x}$ does not belong to any patch, then compute the output $y(\mathbf{x})$ using the updated global fuzzy system.
\end{itemize}

\subsection{Patches in a Fuzzy System}

Identifying the appropriate patches is a challenging task. When a fuzzy system is used to construct the initial global model, we propose to use the first-order rule partitions \cite{Mendel2018} as the patch candidates, and then select those with the largest sum of squared errors (SSE) from them as the patches.

\begin{figure}[htbp]         \centering
\includegraphics[width=.9\linewidth,clip]{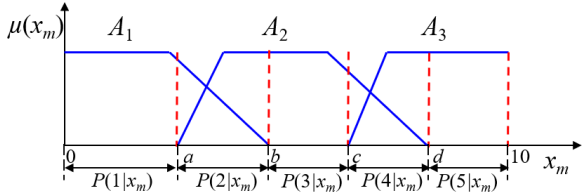}
\caption{First-order rule partitions for $x_m$.} \label{fig:Partition}
\end{figure}

According to \cite{Mendel2018}, first-order rule partitions for $x_m$ in a fuzzy system are a collection of non-overlapping intervals in $D_m$, in each of which the same number of same rules is fired whose firing levels contribute to the output of that system. For example, Fig.~\ref{fig:Partition} shows the five first-order rule partitions for $x_m$, denoted by $P(1|x_m),...,P(5|x_m)$ (more details on how to identify the first-order rule partitions can be found in \cite{Mendel2018}). For a fuzzy system with $M$ inputs, each with $K_m$ ($m=1,...,M$) first-order rule partitions, the total number of first-order rule partitions is $K=\prod_{m=1}^M K_m$, i.e., the total number of all different combinations of the first-order rule partitions in different input domains.

For a well-optimized fuzzy system, transition from one rule partition to another changes the functional form of the input-output mapping. For example, in Fig.~\ref{fig:Partition}, assume $x_m$ is the only input of the TSK fuzzy system, which has the following three rules:
\begin{align*}
R_1: &\mbox{ IF } x_m \mbox{ is } A_1, \mbox{ THEN } y=y_1(x_m)\\
R_2: &\mbox{ IF } x_m \mbox{ is } A_2, \mbox{ THEN } y=y_2(x_m)\\
R_3: &\mbox{ IF } x_m \mbox{ is } A_3, \mbox{ THEN } y=y_3(x_m)
\end{align*}
where $y_1(x_m)$, $y_2(x_m)$ and $y_3(x_m)$ are different functions of $x_m$. In Partition $P(1|x_m)$, only Rule $R_1$ is fired, and hence the fuzzy system output is $y=f_1(x_m)$; in Partition $P(2|x_m)$, both Rules $R_1$ and $R_2$ are fired, and hence the fuzzy system output is:
\begin{align}
y=\frac{\mu_{A_1}(x_m)y_1(x_m)+\mu_{A_2}(x_m)y_2(x_m)}{\mu_{A_1}(x_m)+\mu_{A_2}(x_m)};
\end{align}
in Partition $P(3|x_m)$, only Rule $R_2$ is fired, and hence the fuzzy system output is $y=f_2(x_m)$. Clearly, the functional forms of $y$ in the different rule partitions are different. It is reasonable to believe this is because the (unknown) groundtruth functional form changes significantly from one rule partition to another, and hence the fuzzy system uses different functional forms in different rule partitions to accommodate it. Thus, we can consider each such partition as a patch candidate\footnote{This is just one simple way to initialize the patch candidates. There may be better ways to do this.}.

According to Pedrycz \cite{Pedrycz2018}, \emph{``by information granules one regards a collection of elements drawn together by their closeness (resemblance, proximity, functionality, etc.) articulated in terms of some useful spatial, temporal, or functional relationships. Subsequently, Granular Computing is about representing, constructing, processing, and communicating information granules."} All points within a first-order rule partition fire the same number of same rules, and hence they resemble each other. So, each first-order rule partition can be viewed as an information granule, and PL can be viewed as a form of granular computing.

\subsection{Implementation Details}

For the ease of programming implementation, we can use a single index $k\in[1,K]$ to denote a patch candidate (first-order rule partition) $(P(k_1|x_1),P(k_2|x_2),...,P(k_M|x_M))$, where
\begin{align}
k&=(k_1-1)\cdot \prod_{m=2}^M K_m+(k_2-1)\cdot \prod_{m=3}^M K_m \nonumber \\
 &\quad +\cdots+(k_{M-1}-1)\cdot K_M+k_M\nonumber \\
&=k_M+\sum_{m=1}^{M-1}\left[ (k_m-1)\cdot \prod_{p=m+1}^M K_p\right] \label{eq:k}
\end{align}
For example, assume the fuzzy system has two inputs ($M=2$), $x_1$ and $x_2$, each with five first-order rule partitions ($K_1=K_2=5$) shown in Fig.~\ref{fig:Partition}. Then, (\ref{eq:k}) becomes $k=(k_1-1)\times K_2+k_2$. The partition $(P(1|x_1),P(1|x_2))$ is mapped to $k=(1-1)\times 5+1=1$, $(P(1|x_1),P(2|x_2))$ to $k=(1-1)\times 5+2=2$, $\cdots$, $(P(2|x_1),P(1|x_2))$ to $k=(2-1)\times5+1=6$, $\cdots$, and $(P(5|x_1),P(5|x_2))$ to $k=(5-1)\times 5+5=25$.

Because the first-order rule partitions for $n$ variables begin by examining those partitions for each variable separately, the geometry (dimensions) of each partition is easily known. For example, $(P(4|x_1), P(5|x_2))$ is $[c,d]\times[d,10]$ in Fig.~\ref{fig:Partition}.

For a given $k\in[1,K]$, we can also map it back to a first-order rule partition $(P(k_1|x_1),P(k_2|x_2),...,P(k_M|x_M))$:
\begin{align}
k_1&=int\left(\frac{k-1}{\prod_{m=2}^M K_m}\right)+1 \label{eq:k1}\\
k_2&=int\left(\frac{k-1-(k_1-1)\cdot \prod_{m=2}^M K_m}{\prod_{m=3}^M K_m}\right)+1\\
k_m&=int\left(\frac{k-1-\sum_{i=1}^{m-1}\left[(k_i-1)\cdot \prod_{p=i+1}^M K_p\right]}{\prod_{p=m+1}^M K_p}\right)+1\\
k_M&=k-\sum_{m=1}^{M-1}\left[ (k_m-1)\cdot \prod_{p=m+1}^M K_p\right] \label{eq:kM}
\end{align}
where $int(x)$ means the integer part of $x$, e.g., $int(2.0)=2$ and $int(2.9)=2$. Using again the above example, when $k=6$, we have $k_1=int(\frac{k-1}{K_2})+1=int(\frac{6-1}{5})+1=2$ and $k_2=k-(k_1-1)\times K_2=6-(2-1)\times 5=1$, and hence the mapped first-order rule partition for $k=6$ is $(P(2|x_1),P(1|x_2))$.

In summary, the pseudo-code\footnote{A sample Matlab implementation is available at https://github.com/drwuHUST/Patch-Learning.} in Algorithm~\ref{alg:PLFS} implements the three generic PL steps described at the beginning of Section~\ref{sect:PL} using fuzzy systems. Its limitations and potential improvements are discussed in Section~\ref{sect:limitations}, after some experiments to demonstrate its capability are described in the next section.

\afterpage{\begin{algorithm}[htbp] %\DontPrintSemicolon
\KwIn{$N$ labeled training examples, $\{(\mathbf{x}_n,y_n)\}^N_{n=1}$, where $\mathbf{x}_n\in\mathbb{R}^{M\times 1}$\;
\hspace*{9mm} $T$ unlabeled examples, $\{\mathbf{x}_t\}_{t=1}^T$\;
\hspace*{9mm} $L$, the maximum number of patch models to be trained\;}
\KwOut{The PL model predictions for $\{\mathbf{x}_t\}_{t=1}^T$.}
\tcp{Train the PL model}
Train a global fuzzy model using all $N$ training examples\;
\For{$m=1,...,M$}
{Identify the first-order rule partitions for the $m$th input domain of the global fuzzy model\;}
Index the partitions using $k$ in (\ref{eq:k})\;
Include all partitions in the candidate pool\;
$l=1$\;
    \While{$l\le L$}{
    Identify from the candidate pool the partition giving the maximum SSE\;
    Record the location of the partition as the $l$th patch\;
    Train a patch fuzzy model using only the training examples within the $l$th patch\;
    \If{the $l$th model is successfully trained\footnotemark{}}
    {$l=l+1$\;}
    Remove the above patch from the candidate pool\;}
Update the global fuzzy model, using only the training examples that do not fall into any patch\;
\tcp{Use the PL model for prediction}
    \For{$t=1,...,T$}{
    $useGlobal=1$\;
    \For{$l=1,...,L$}{
    \If{$\mathbf{x}_t$ falls into the $l$th patch}{
        Predict using the $l$th patch fuzzy model\;
        $useGlobal=0$\;
        Break\;}}
        \If{$useGlobal==1$}{
        Predict using the updated global fuzzy model\;}}
\caption{PL using fuzzy systems.} \label{alg:PLFS}
\end{algorithm}
\footnotetext{A fuzzy model may not be trainable for a patch, e.g., where there are too few training examples within that patch.}}

\section{Experimental Results} \label{sect:experiments}

This section demonstrates our proposed fuzzy systems-based PL. All global models and patch models were trained by adaptive-network-based fuzzy inference system (ANFIS) \cite{Jang1993} using trapezoidal input MFs and first-order TSK rule consequents. For example, when there are two inputs, the global model and patch model rules are in the form of:
\begin{align*}
\mbox{IF } x_1 \mbox{ is } A_1 \mbox{ and } x_2 \mbox{ is } A_2, \mbox{ THEN } y=b_0+b_1x_1+b_2x_2
\end{align*}
where $A_1$ and $A_2$ are trapezoidal fuzzy sets, and $b_0$, $b_1$ and $b_2$ are adjustable coefficients.

We also compared PL with two classic ensemble learning approaches for regression, briefly described in the Introduction:
\begin{enumerate}
\item Bagging \cite{Breiman1996}, which bootstraps (samples with replacement) the training set repeatedly, trains an ANFIS model from each replicate, and then aggregates all models by averaging. Bagging mainly reduces the variance of the estimation.
\item LSBoost (least squares boost) \cite{Drucker1997}, which is a boosting algorithm for regression problems. We tried to also use ANFIS as the weak learner here. Unfortunately, LSBoost requires the weak learners to be able to use weighted training examples, but we are not aware of such an ANFIS algorithm\footnote{We could develop such an ANFIS algorithm, but it is beyond the scope of this paper. Additionally, this demonstrates one advantage of PL: it does not add weights to the training examples, so any existing machine learning algorithms can be readily used as the patch model.}. So, we used regression tree as the weak learner (\emph{fitrensemble} in Matlab), which is also the most popular choice in boosting algorithms. Boosting mainly reduces the bias of the estimation.
\end{enumerate}
The number of weak learners in Bagging and LSBoost was set to the number of total models in PL, i.e., $L+1$, so that a fair comparison can be made.

\subsection{Experiment 1: 1D Curve Fitting}

The first experiment is the same as the simple example introduced in Section~\ref{sect:PL}. It was chosen because we can easily visualize the patch models and intermediate results for this simple 1D curve fitting experiment. Two input MFs were used in the ANFIS models.

\begin{figure}[htbp]\centering
\subfigure[]{\label{fig:f1PL0}   \includegraphics[width=.47\linewidth,clip]{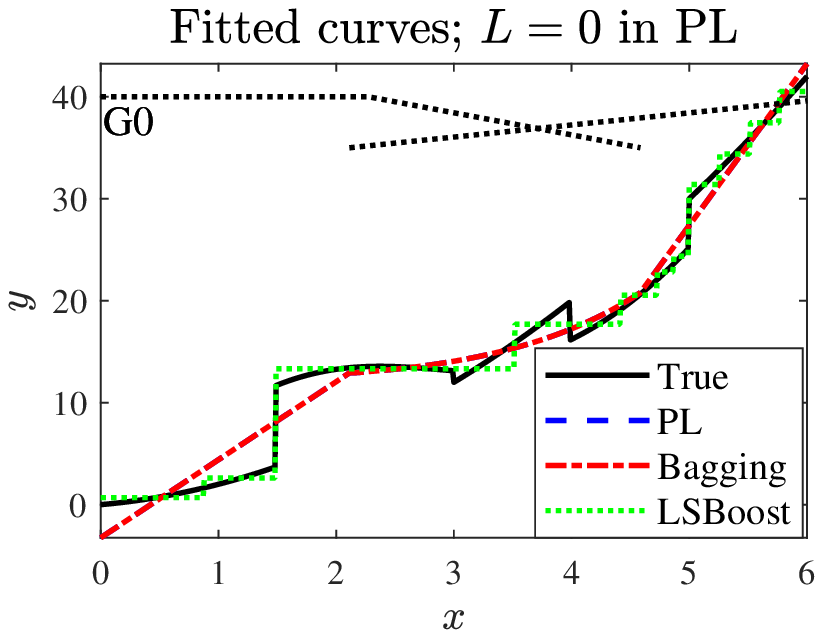}}
\subfigure[]{\label{fig:f1RMSE0}    \includegraphics[width=.47\linewidth,clip]{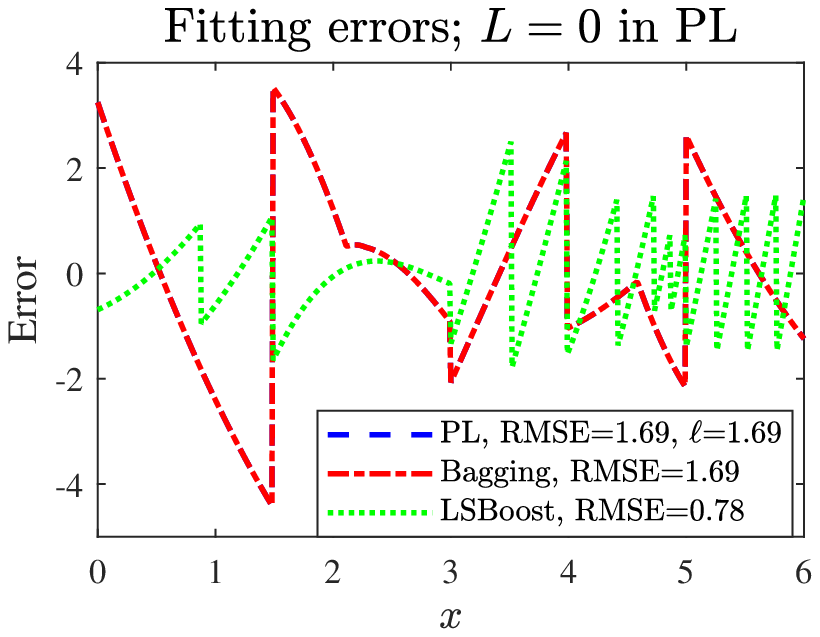}}
\subfigure[]{\label{fig:f1PL1}   \includegraphics[width=.47\linewidth,clip]{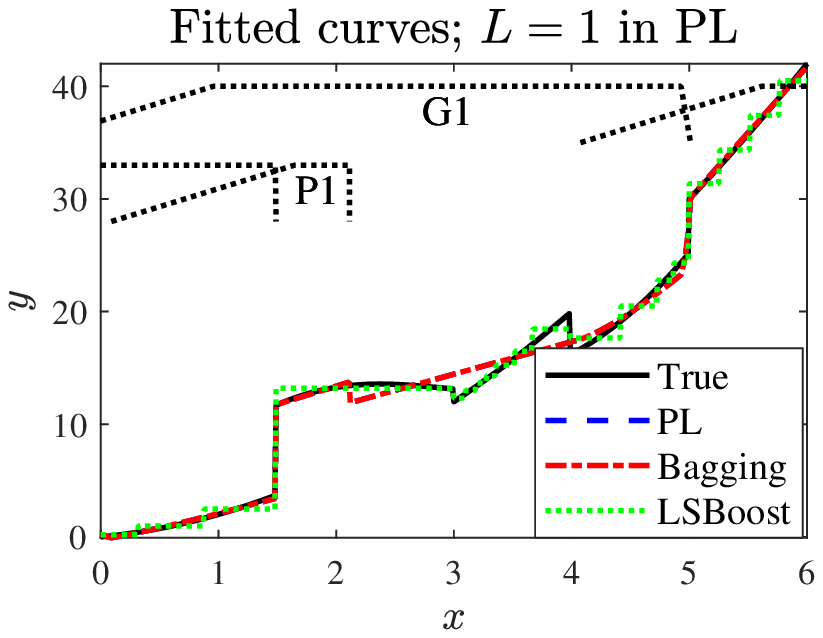}}
\subfigure[]{\label{fig:f1RMSE1}    \includegraphics[width=.47\linewidth,clip]{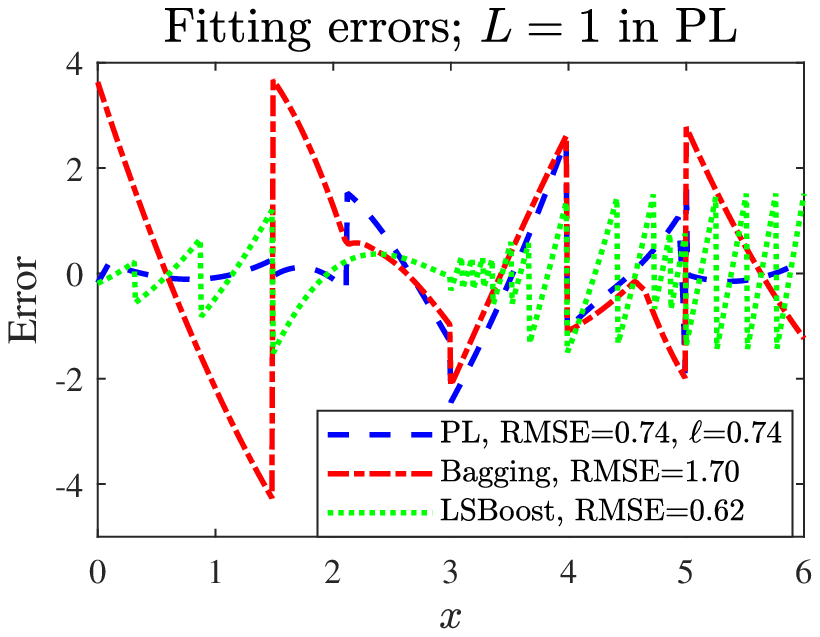}}
\subfigure[]{\label{fig:f1PL2}   \includegraphics[width=.47\linewidth,clip]{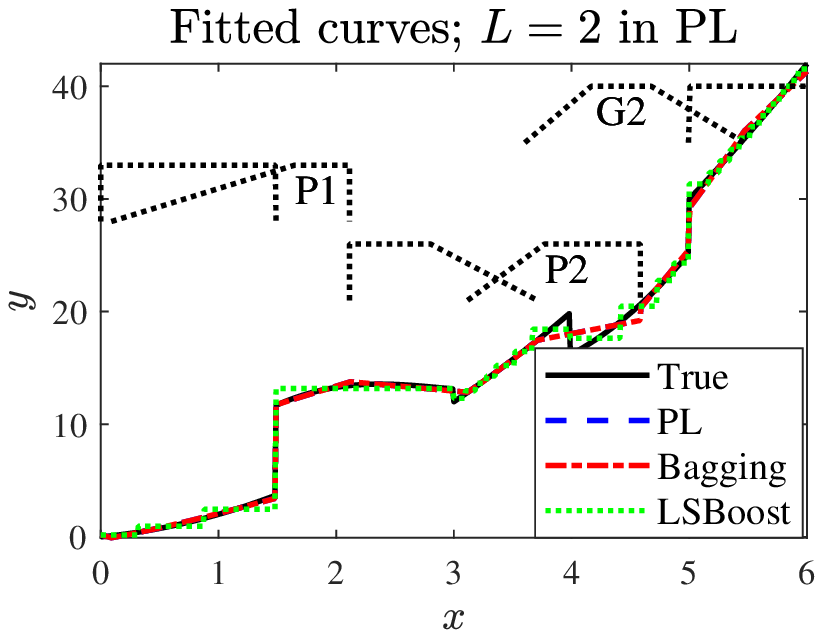}}
\subfigure[]{\label{fig:f1RMSE2}    \includegraphics[width=.47\linewidth,clip]{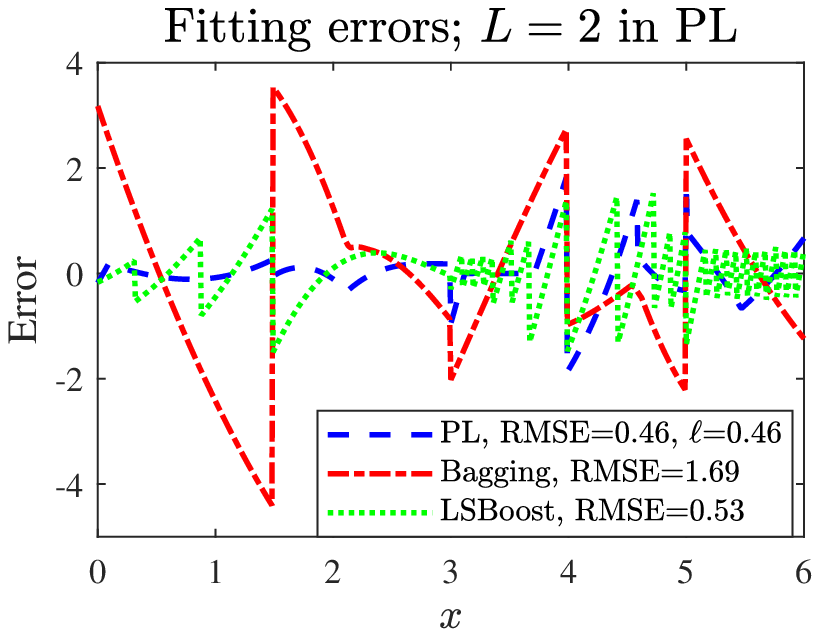}}
\caption{Experiment~1: 1D curve fitting using ANFIS-based PL (with two MFs in the input domain), Bagging and LSBoost. (a) fitting with $L=0$ in PL, and one weak learner in Bagging and LSBoost; (b) the corresponding fitting errors; (c) fitting with $L=1$ in PL, and two weak learners in Bagging and LSBoost; (d) the corresponding fitting errors; (e) fitting with $L=2$ in PL, and three weak learners in Bagging and LSBoost; (f) the corresponding fitting error.} \label{fig:f1PL}
\end{figure}

When only the global fuzzy model was used to fit the 1D curve in (\ref{eq:g}), the fitted curve is shown in Fig.~\ref{fig:f1PL0}, and the corresponding errors are shown in Fig.~\ref{fig:f1RMSE0}. The two trapezoidal input MFs in the global fuzzy model are shown at the top of Fig.~\ref{fig:f1PL0} as the black dotted curves (around the label `G0'). The corresponding first-order rule partitions (patch candidates) were $[0, 2.11]$, $[2.11,4.58]$ and $[4.58,10]$, and their SSEs were $1230.8$, $250.2$ and $234.4$, respectively.

Since the RMSE using only one global fuzzy model was too large (which was $1.69$, corresponding to $\ell=1.69$), we identified the first patch as $[0,2.11]$, used the 212 training examples falling into it to train the corresponding patch fuzzy model, and then updated the global fuzzy model. The new fitting is shown in Fig.~\ref{fig:f1PL1}, and the corresponding input trapezoidal MFs for the global fuzzy model and patch fuzzy model are shown at the top of the same figure, as the black dotted curves (the updated global model is around the label `G1'; the patch model is around the label `P1'). The patch fuzzy model focused on the interval $[0,2.11]$, and reduced the RMSE\footnote{The original SSE in $[0,2.11]$ was $1230.8$, and there were 212 points; hence, the original RMSE was $\sqrt{1230.8/212}=2.41$.} in this interval from $2.41$ to $0.11$, as shown in Fig.~\ref{fig:f1RMSE1}. Note that the global fuzzy model was updated after adding the patch fuzzy model, so its MFs were different from those in Fig.~\ref{fig:f1PL0}. The patch fuzzy model and the updated global fuzzy model together reduced the overall RMSE from $1.69$ to $0.74$, and the loss $\ell$ from $1.69$ to $0.88$, a significant improvement.

We can proceed to add another patch fuzzy model for $[2.11,4.58]$ to further reduce the fitting error. The results are shown in Figs.~\ref{fig:f1PL2} and \ref{fig:f1RMSE2}, and the corresponding input trapezoidal MFs for the global fuzzy model and patch fuzzy models are shown at the top of Fig.~\ref{fig:f1PL2}. The initial global fuzzy model was the same as the one in Fig.~\ref{fig:f1PL0}, and the three first-order partitions were still $[0, 2.11]$, $[2.11,4.58]$ and $[4.58,10]$. The first patch fuzzy model, whose input MFs are represented as the black dotted curves around the label `P1' at the top of Fig.~\ref{fig:f1PL2}, still focused on the interval $[0,2.11]$, identical to the patch fuzzy model in Fig.~\ref{fig:f1PL1}. The second patch fuzzy model focused on the interval $[2.11,4.58]$, and its input MFs are represented as the black dotted curves around the label `P2' at the top of Fig.~\ref{fig:f1PL2}. With the help of the second patch fuzzy model, the RMSE\footnote{The original SSE in $[2.11,4.58]$ was $250.2$, and there were 247 points; hence, the original RMSE was $\sqrt{250.2/247}=1.01$.} in $[2.11,4.58]$ was reduced from $1.01$ to $0.65$. Finally, the input MFs of the updated global fuzzy model are shown as the black dotted curves around the label `G2' at the top of Fig.~\ref{fig:f1PL2}. Its MFs covered the interval $[3.60,6]$ instead of the full range $[0,6]$, because the range $[0,4.58]$ had been fully covered by the two patch fuzzy models, and hence the updated global fuzzy model only needed to ensure the interval $(4.58,6]$ was covered. The two patch fuzzy models and the updated global fuzzy model together reduced the overall RMSE to $0.46$, and the loss $\ell$ to $0.60$, a significant improvement over using zero or one patch fuzzy model.

At this point we had only one patch candidate left, which was $[4.58,6]$. We should not add more patch models, because the final patch candidate will be handled by the default global model (if we add another patch model for this interval, then there will be no training data left for the default global model). As we achieved the smallest loss $\ell$ when two patch models were used, our final PL model should include two patches, as shown in Fig.~\ref{fig:f1PL2}.

It's also interesting to compare the performance of PL with those of Bagging and LSBoost. Bagging and PL had identical performance when only the global model was used. However, unlike PL, whose performance improved as patch models were added, the performance of Bagging did not change with the number of models. Interestingly, although at the beginning LSBoost had smaller RMSE than PL (when only the global model was used), when two patch models were added, PL outperformed LSBoost in terms of RMSE.

\subsection{Experiment 2: 2D Surface Fitting}

\begin{figure}[htbp]         \centering
\includegraphics[width=.8\linewidth,clip]{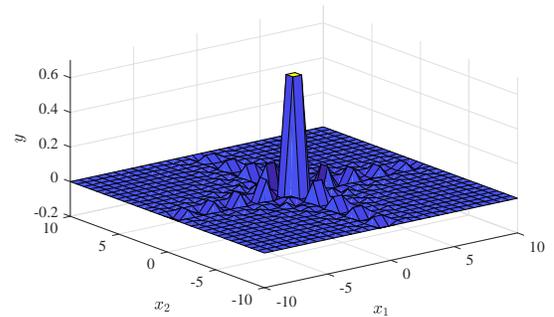}
\caption{The 2D surface in Experiment~2.} \label{fig:Y2}
\end{figure}

The second experiment, which is Example~1 in \cite{Jang1993}, considered a more complex regression problem: 2D curve fitting. The 2D surface was
\begin{align}
y=\frac{\sin(x_1)}{x_1}\times \frac{\sin(x_2)}{x_2}, \quad x_1,x_2\in[-10,10],
\end{align}
as shown in Fig.~\ref{fig:Y2}. 30 uniform samples were used for both $x_1$ and $x_2$. Hence, there were a total of 900 training examples.

\begin{figure}[htbp]\centering
\includegraphics[width=.8\linewidth,clip]{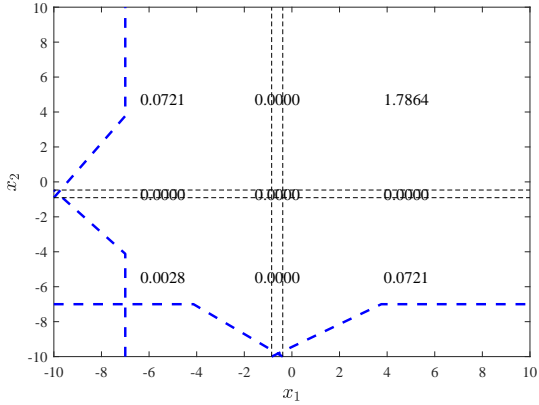}
\caption{The MFs, first-order rule partitions, and their corresponding SSE in Experiment~2.} \label{fig:FP2}
\end{figure}

\begin{figure*}[htbp]\centering
\subfigure[]{\label{fig:f2PL0}   \includegraphics[width=.23\linewidth,clip]{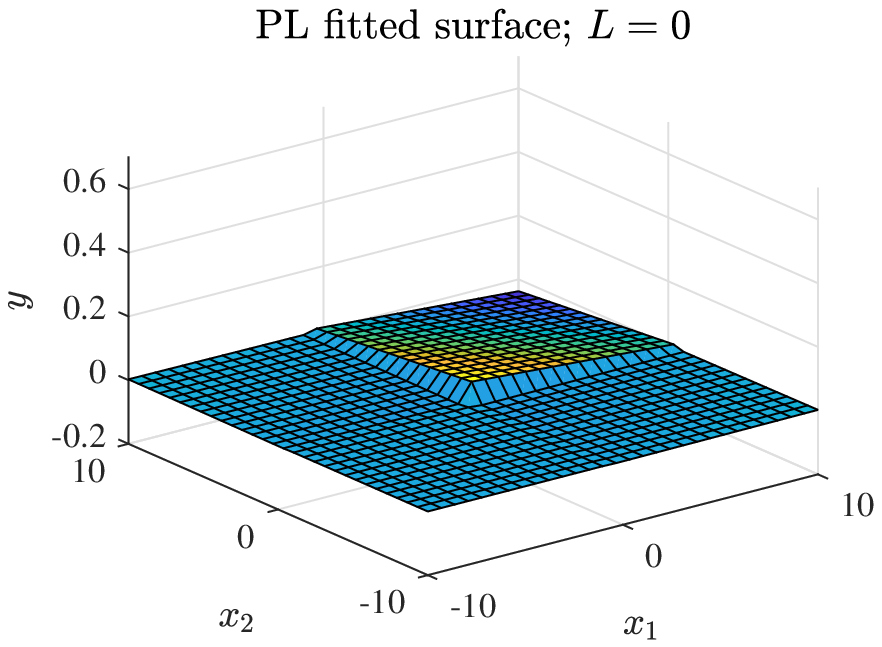}}
\subfigure[]{\label{fig:f2RMSE0}    \includegraphics[width=.23\linewidth,clip]{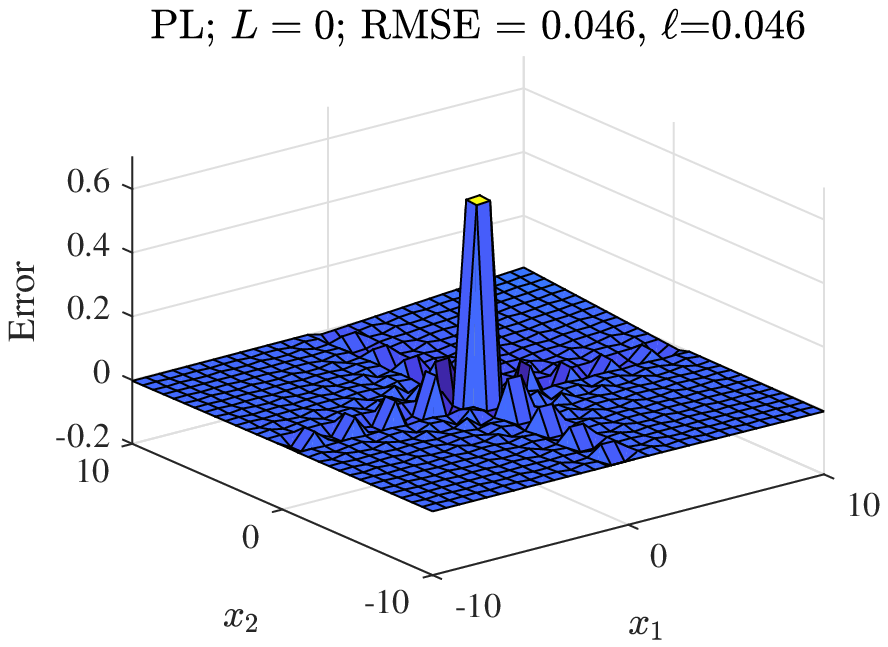}}
\subfigure[]{\label{fig:f2Bagging1}   \includegraphics[width=.23\linewidth,clip]{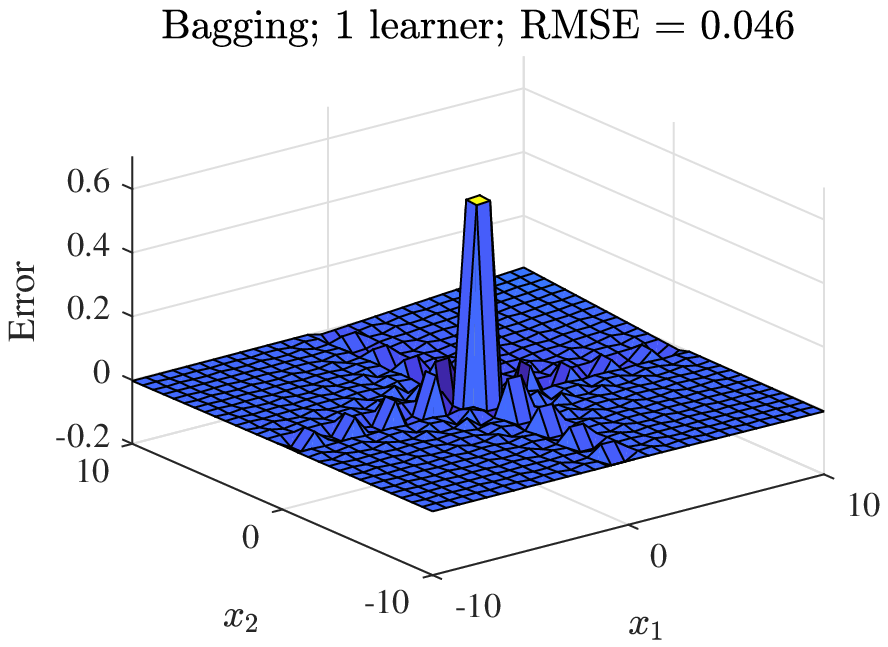}}
\subfigure[]{\label{fig:f2LSBoost1}    \includegraphics[width=.23\linewidth,clip]{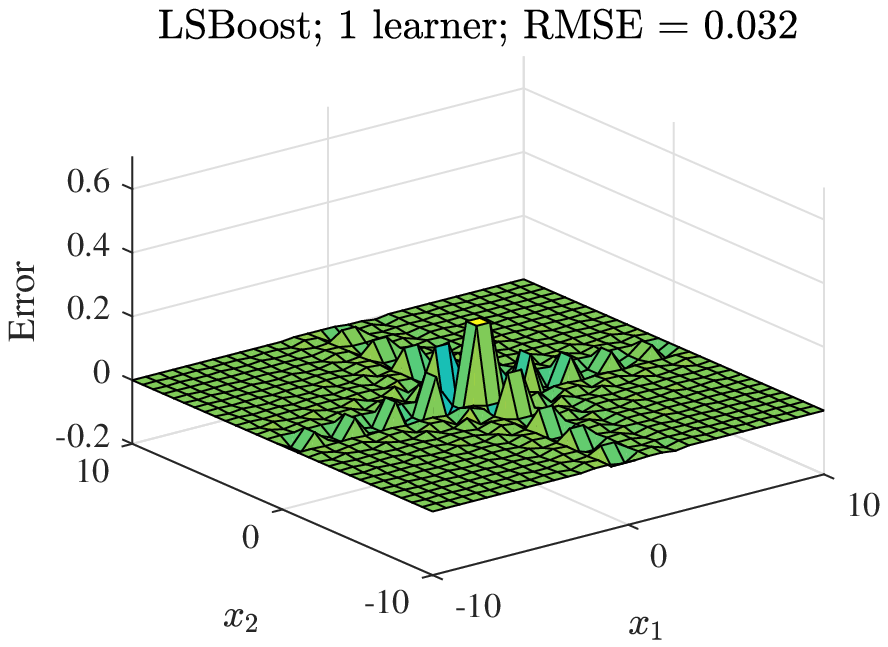}}
\subfigure[]{\label{fig:f2PL1}   \includegraphics[width=.23\linewidth,clip]{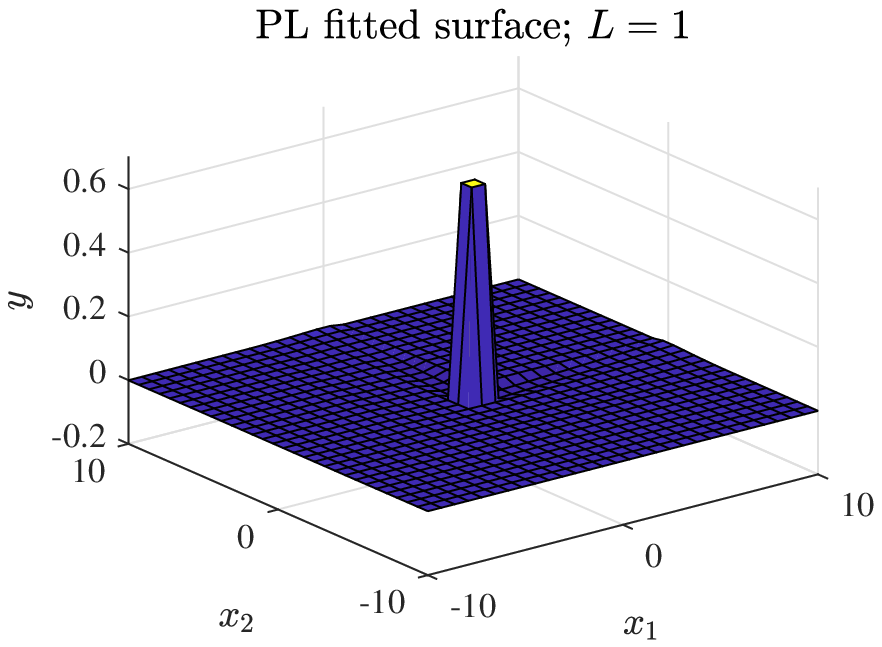}}
\subfigure[]{\label{fig:f2RMSE1}    \includegraphics[width=.23\linewidth,clip]{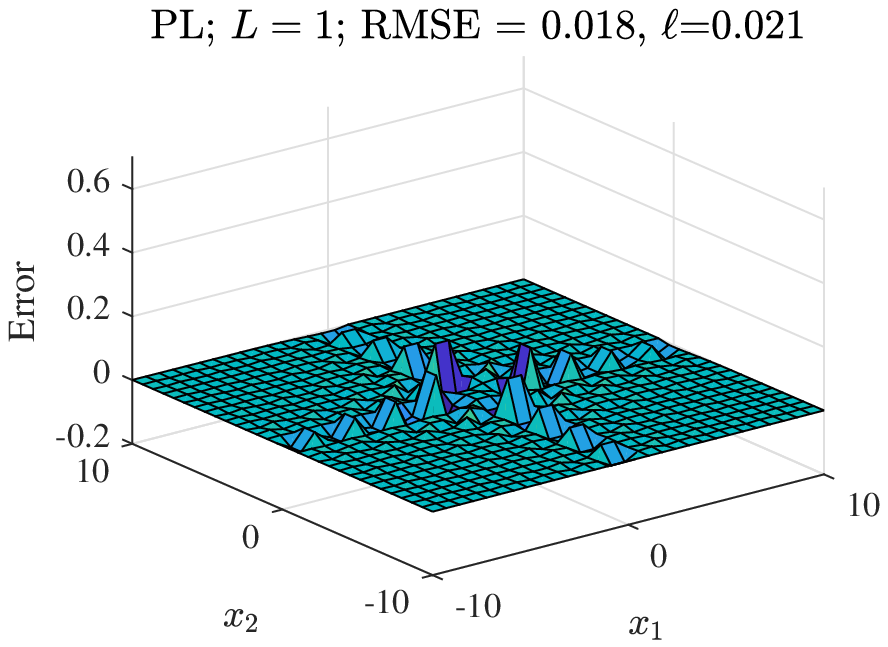}}
\subfigure[]{\label{fig:f2Bagging2}   \includegraphics[width=.23\linewidth,clip]{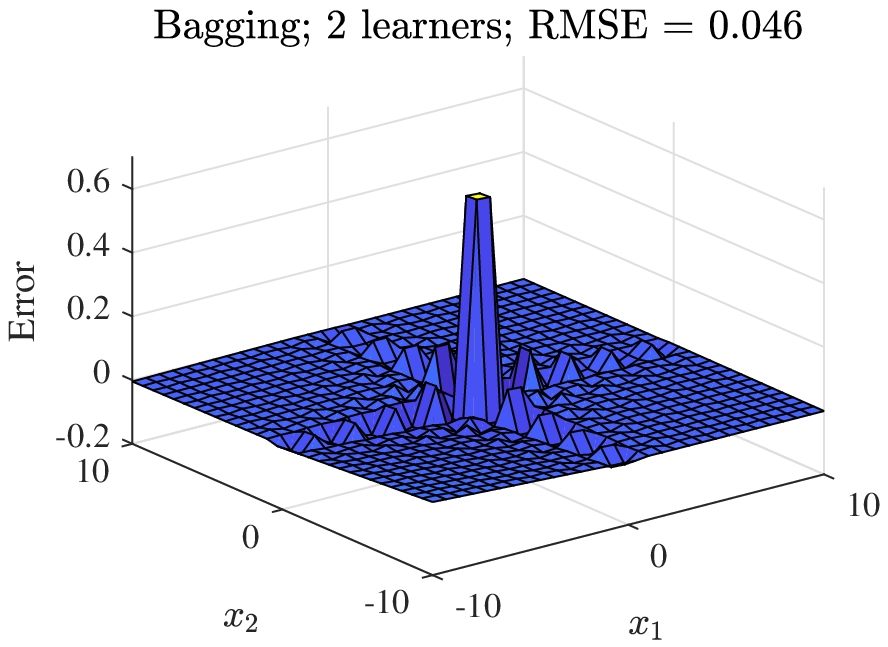}}
\subfigure[]{\label{fig:f2LSBoost2}    \includegraphics[width=.23\linewidth,clip]{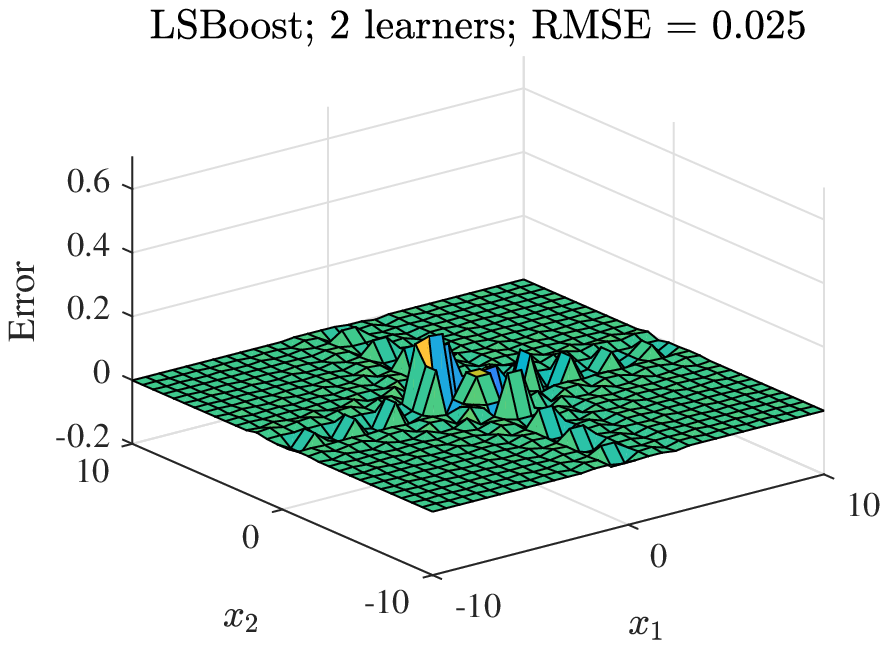}}
\subfigure[]{\label{fig:f2PL2}   \includegraphics[width=.23\linewidth,clip]{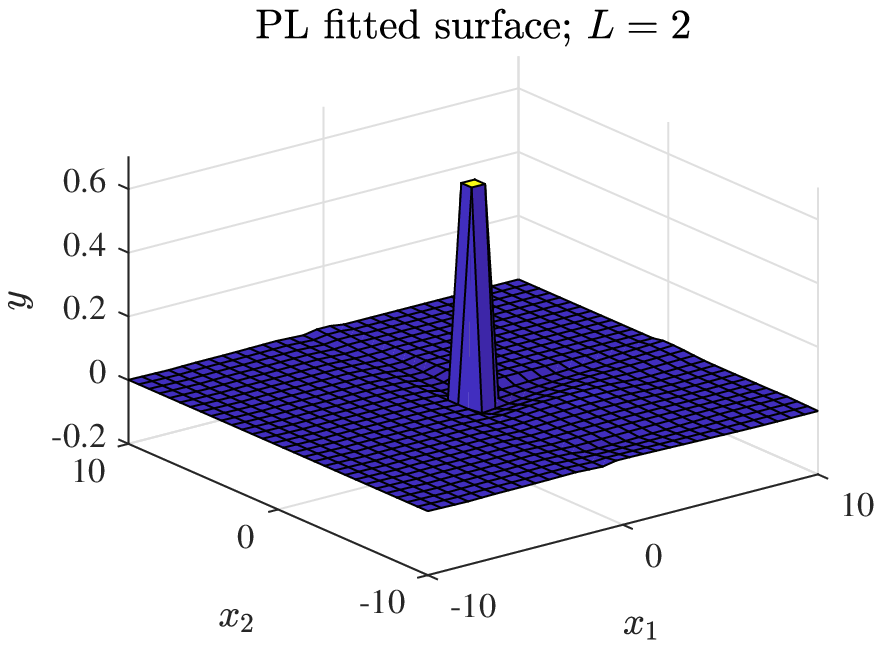}}
\subfigure[]{\label{fig:f2RMSE2}    \includegraphics[width=.23\linewidth,clip]{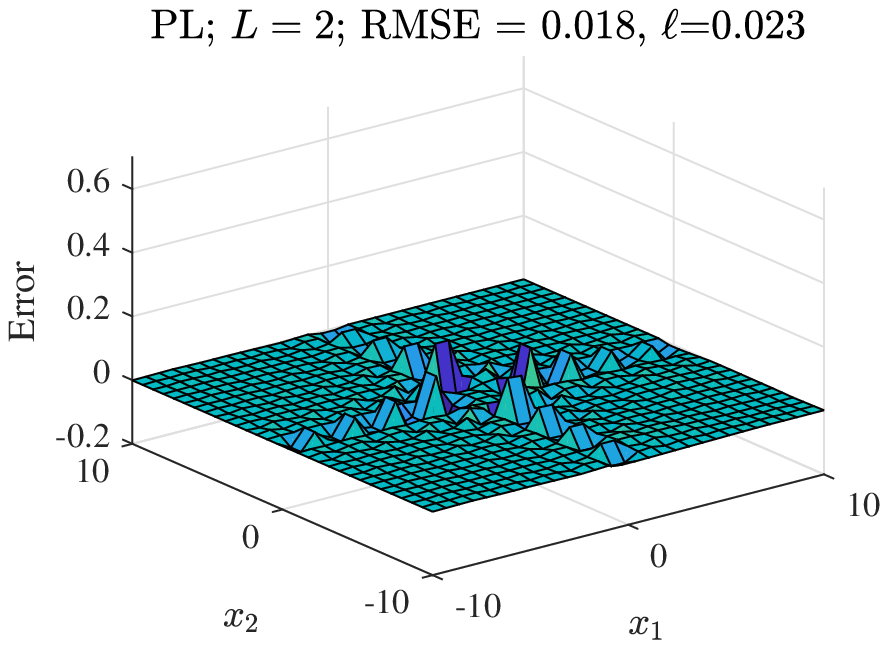}}
\subfigure[]{\label{fig:f2Bagging3}   \includegraphics[width=.23\linewidth,clip]{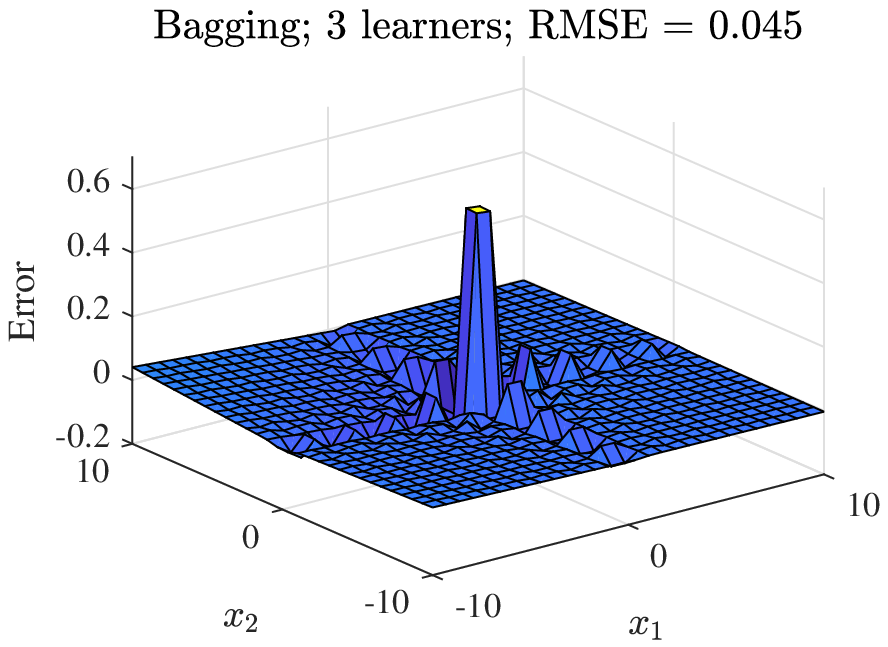}}
\subfigure[]{\label{fig:f2LSBoost3}    \includegraphics[width=.23\linewidth,clip]{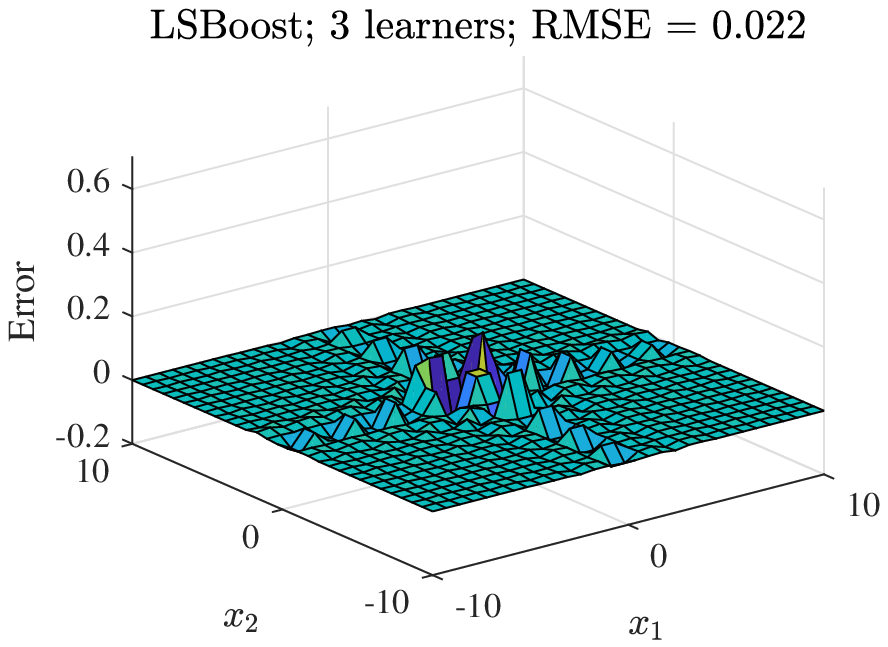}}
\caption{Experiment~2: 2D surface fitting using ANFIS-based PL, Bagging, and LSBoost. (a) fitted surface by PL, $L=0$; (b) the corresponding fitting error of PL; (c) fitting error of Bagging, with one weak learner; (d) fitting error of LSBoost, with one weak learner; (e) fitted surface by PL, $L=1$; (f) the corresponding fitting error of PL; (g) fitting error of Bagging, with two weak learners; (h) fitting error of LSBoost, with two weak learners; (i) fitted surface by PL, $L=2$; (j) the corresponding fitting error of PL; (k) fitting error of Bagging, with three weak learners; (l) fitting error of LSBoost, with three weak learners.} \label{fig:f2PL}
\end{figure*}

We used ANFIS with two trapezoidal MFs [shown as the blue dashed curves in Fig.~\ref{fig:FP2}] in each input domain as the global and patch fuzzy models. The performances of the global fuzzy model are shown in Figs.~\ref{fig:f2PL0} and \ref{fig:f2RMSE0}. Observe that a single global fuzzy model cannot fit the 2D surface well, and resulted in a large RMSE. Fig.~\ref{fig:FP2} also shows the nine first-order rule partitions (patch candidates) and their corresponding SSEs. The five partitions around $x_1=0$ or $x_2=0$ had zero SSE, because they were so narrow that no training examples fell into them.

When a patch fuzzy model was added to the partition $[-0.3845,10]\times[-0.4658,10]$ (whose SSE was $1.7864$) in Fig.~\ref{fig:FP2} and the global fuzzy model was updated, the results are shown in Figs.~\ref{fig:f2PL1} and \ref{fig:f2RMSE1}. The fitting looked much more similar to the groundtruth in Fig.~\ref{fig:Y2}. The RMSE was reduced from 0.046 to 0.018, and the loss was reduced from 0.046 to 0.021.

When a second patch fuzzy model was added to the partition $[-0.3845,10]\times[-10,-0.9038]$  (whose SSE was $0.0721$) in Fig.~\ref{fig:FP2} and the global fuzzy model was updated, the results are shown in Figs.~\ref{fig:f2PL2} and \ref{fig:f2RMSE2}. The RMSE did not change, whereas the loss increased from 0.021 to 0.023, suggesting that adding the second patch fuzzy model was not beneficial. So, the final PL model should include only one patch.

The fitting errors of Bagging using one, two and three weak learners are shown in Figs.~\ref{fig:f2Bagging1}, \ref{fig:f2Bagging2} and \ref{fig:f2Bagging3}, respectively. The fitting errors of LSBoost using one, two and three weak learners are shown in Figs.~\ref{fig:f2LSBoost1}, \ref{fig:f2LSBoost2} and \ref{fig:f2LSBoost3}, respectively. Again, the performance of Bagging did not change with the number of weak learners. Although LSBoost outperformed PL when only the global model was used, PL outperformed LSBoost when a patch fuzzy model was added.

\subsection{Experiment 3: 3D Manifold Fitting}

Experiment~3, which is Example~2 in \cite{Jang1993}, considered an even more complex regression problem: fitting a 3D manifold
\begin{align}
y=(1+x_1^{0.5}+x_2^{-1}+x_3^{-1.5})^2, \quad x_1,x_2,x_3\in[1,6]
\end{align}
11 uniform samples were used for each of $x_1$, $x_2$ and $x_3$. Simple ANFISs with two trapezoidal MFs in each input domain were used as the global and patch fuzzy models.

Since there is no easy way to visualize the fitting results and the errors for this manifold, we only show the resulting RMSEs, average percentage errors\footnote{This was the performance measure used in \cite{Jang1993}.} (APEs), and losses in the top panel of Table~\ref{tab:E3}, for different $L$. The five patches were $[2.01,4.89]\times[1.00,4.59]\times[4.74,6.00]$, $[4.89,6.00]\times[1.00,4.59]\times[4.74,6.00]$, $[2.01,4.89]\times[1.00,4.59]\times[1.48,4.74]$, $[4.89,6.00]\times[1.00,4.59]\times[1.48,4.74]$, and $[0.00,2.01]\times[1.00,4.59]\times[4.74,6.00]$, respectively. As $L$ increased from zero to four, both RMSE and APE decreased, suggesting the effectiveness of PL. The RMSE and APE started to increase when $L$ exceeded four.

The change of loss $\ell$ was more interesting: it first increased when $L$ increased from zero to one, then decreased, and then increased again when $L$ changed from four to five. This is because the first patch fuzzy model did not contribute much to the performance improvement itself, but latter ones did. This is somewhat analogous to the concept of local minimum in traditional machine learning. Since the RMSE, APE and loss when using four patch fuzzy models were all much better than those when using only the global model, we concluded that four patch fuzzy models should be used in this example.

\begin{table}[h] \centering \setlength{\tabcolsep}{1.5mm}
\caption{Experiment~3: RMSEs, APEs and losses of PL, Bagging and LSBoost in 3D manifold fitting.}   \label{tab:E3}
\begin{tabular}{c|c|cccccc}   \hline
 & & \multicolumn{6}{c}{$L$, number of patch fuzzy models}\\ \cline{3-8}
& & 0 & 1 & 2 & 3 &4 &5\\ \hline
\multirow{3}{*}{PL}&RMSE & 0.2192 &   0.2053 &   0.1719  &  0.1555 & 0.1179 & 0.1379\\
&APE & 0.0157 &   0.0139 &   0.0113 &   0.0100 & 0.0073 & 0.0077\\
&$\ell$ &0.2192& 0.2442 & 0.2262&0.2199&0.1762&0.2158\\ \hline \hline
 & & \multicolumn{6}{c}{Number of weak learners}\\ \cline{3-8}
&  & 1 & 2 & 3 &4 &5 & 6\\ \hline
\multirow{2}{*}{Bagging}&RMSE & 0.2192 &   0.2291 &   0.1546  &  0.1924 & 0.1863 & 0.2337\\
&APE & 0.0157 &   0.0163 &   0.0111 &   0.0138 & 0.0134 & 0.0168\\ \hline
\multirow{2}{*}{LSBoost}&RMSE & 1.6534 &   1.2820 &   0.9823  &  0.8261 & 0.7361 & 0.6797\\
&APE & 0.0157 &    0.0822 &   0.0633 & 0.0562 & 0.0496 & 0.0459\\ \hline
\end{tabular}
\end{table}

The RMSEs and APEs of Bagging and LSBoost are shown in the bottom panel of Table~\ref{tab:E3}, for different numbers of weak learners. Overall both of them had worse performance than PL, when the same number of models (weak learners) were used.

\subsection{Experiment 4: Online System Identification}

\begin{figure}[htbp]         \centering
\includegraphics[width=.8\linewidth,clip]{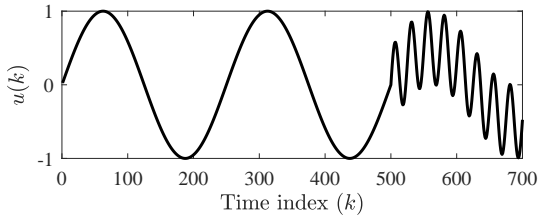}
\caption{$u(k)$ in Experiment~4.} \label{fig:E4u}
\end{figure}

Experiment~4 considers online system identification, which is Example~3 in \cite{Jang1993}. The goal was to use PL to identify a nonlinear component $f(u(k))$ in a control system:
\begin{align}
y(k+1)=0.3y(k)+0.6y(k-1)+f(u(k)),
\end{align}
where $y(k)$ is the output at time index $k$, $u(k)$ is the input:
\begin{align}
u(k)=\left\{\begin{array}{ll}
              \sin\left(\frac{2\pi k}{250}\right), & k\in[1,499] \\
              0.5\sin\left(\frac{2\pi k}{250}\right)+0.5\sin\left(\frac{2\pi k}{25}\right), & k\in[500,700]
            \end{array}\right.
\end{align}
as shown in Fig.~\ref{fig:E4u}, and $f(u)$ is the unknown nonlinear function:
\begin{align}
f(u)=0.6\sin(\pi u)+0.3\sin(3\pi u)+0.1\sin(5\pi u).
\end{align}
PL was used to identify $f(u)$, and its models were updated at each time index $k\in[40,250]$. The PL model updating stopped at $k=250$, and the remaining $700-250=450$ points were used to test the performance of the PL models. Only two trapezoidal MFs were used in the global and patch ANFIS models in PL.

\begin{figure}[htbp]\centering
\subfigure[]{\label{fig:E40}   \includegraphics[width=.8\linewidth,clip]{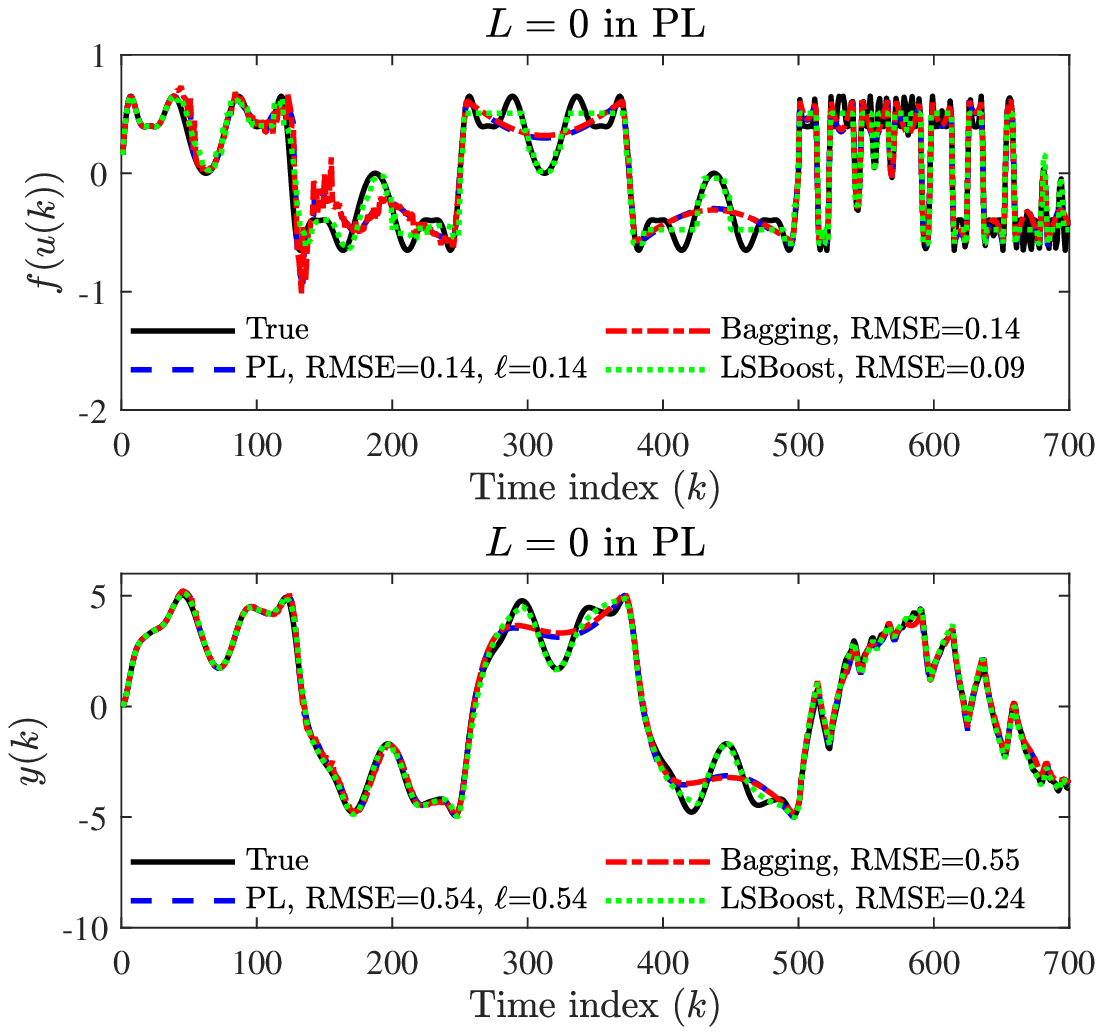}}
\subfigure[]{\label{fig:E41}   \includegraphics[width=.8\linewidth,clip]{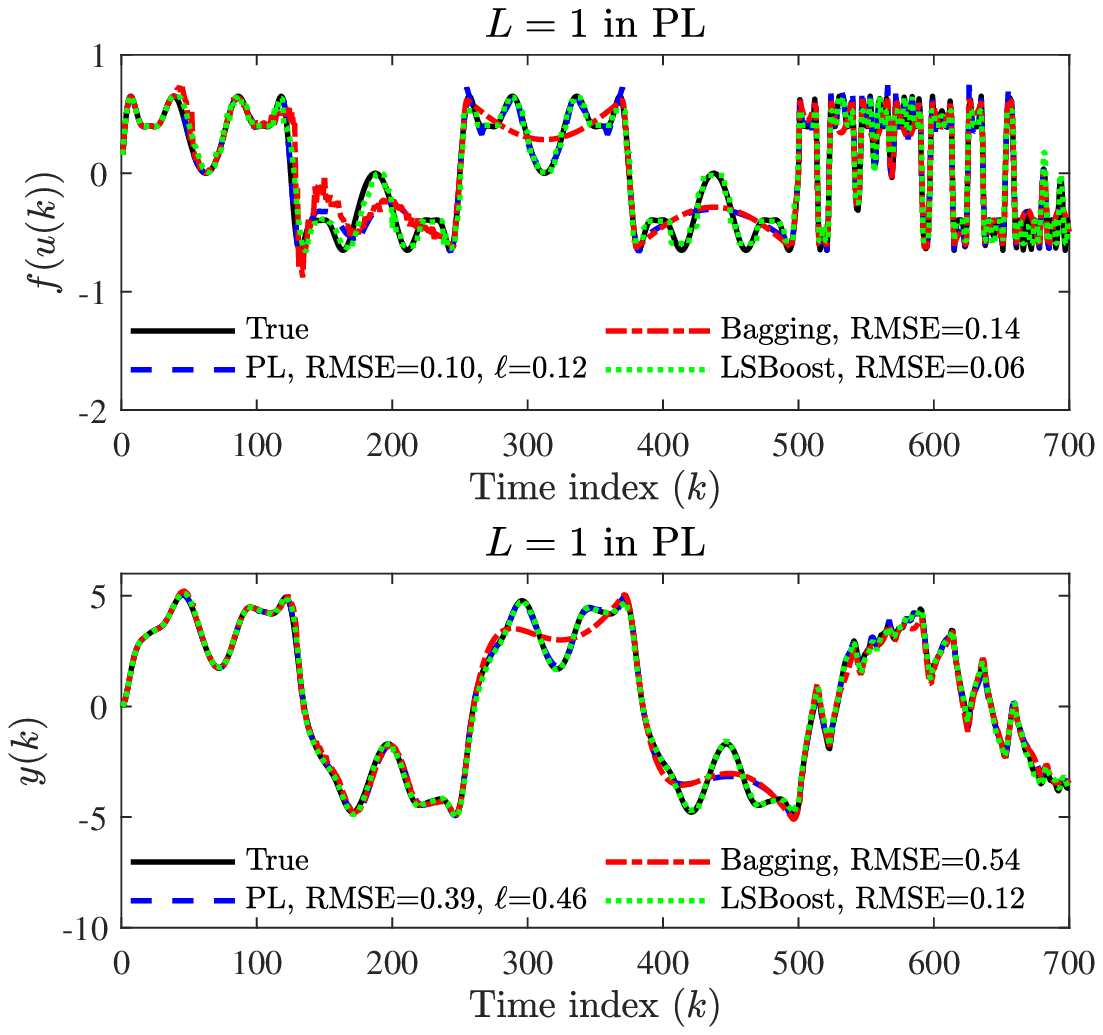}}
\subfigure[]{\label{fig:E42}   \includegraphics[width=.8\linewidth,clip]{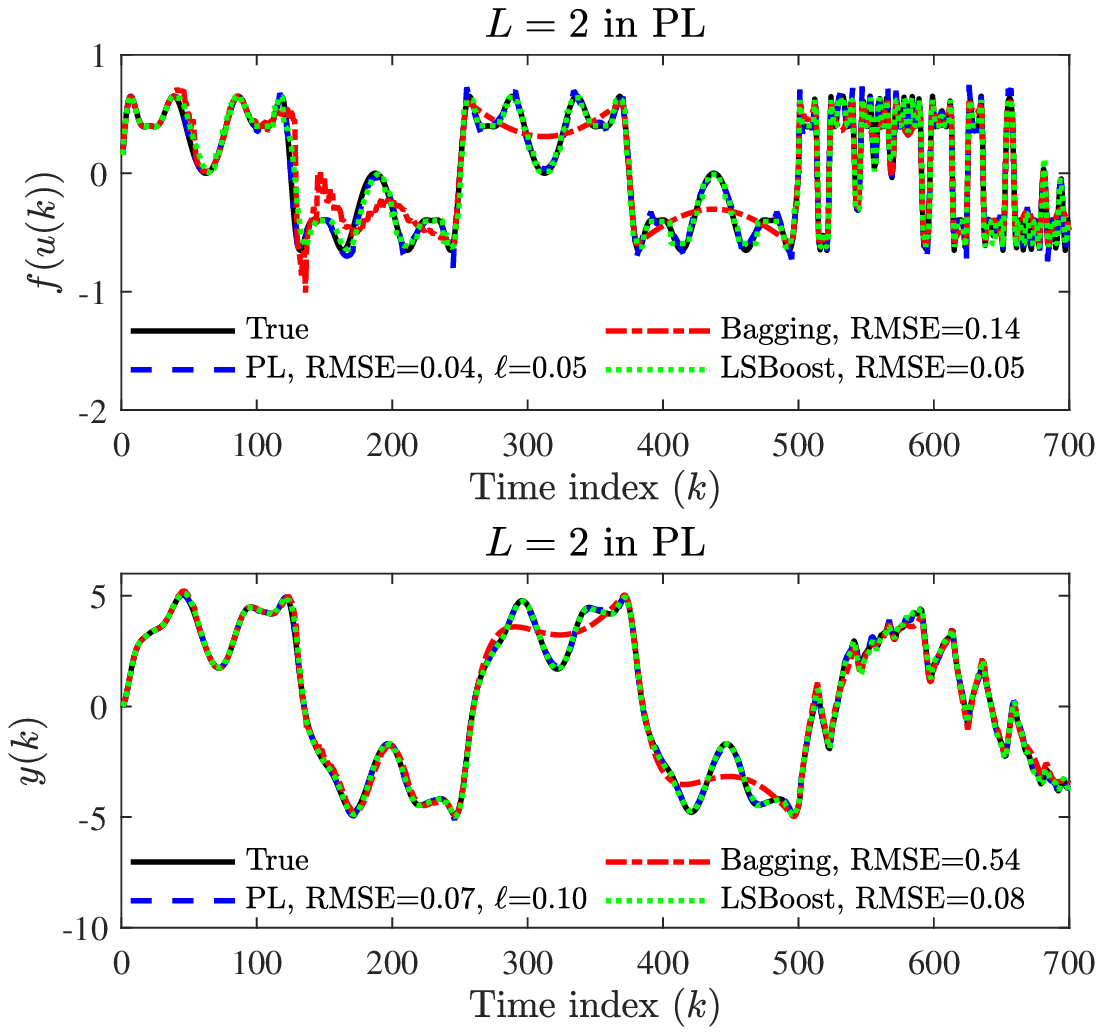}}
\caption{Experiment~4: online system identification using ANFIS-based PL. (a) $f(u(k))$ and $y(k)$ using $L=0$ in PL, and one weak learner in Bagging and LSBoost; (b) $f(u(k))$ and $y(k)$ using $L=1$ in PL, and two weak learners in Bagging and LSBoost; and, (c) $f(u(k))$ and $y(k)$ using $L=2$ in PL, and three weak learners in Bagging and LSBoost.} \label{fig:E4}
\end{figure}

Fig.~\ref{fig:E40} shows the identification results when only the global fuzzy model was used. The titles show the RMSEs on $f(u(k))$ and $y(k)$, computed for $k\in[251,700]$. There were large errors in $f(u(k))$, and hence $y(k)$, i.e., a single 2-MF fuzzy model cannot satisfactorily identify the nonlinear system $f(u(k))$.

Fig.~\ref{fig:E41} shows the identification results when one patch fuzzy model at $[0.11,2.40]$ and the updated global fuzzy model were used. The RMSEs on $f(u(k))$ and $y(k)$ were much smaller than their counterparts in Fig.~\ref{fig:E40}, suggesting that adding a patch fuzzy model was very beneficial.

Fig.~\ref{fig:E42} shows the identification results when two patch fuzzy models (at $[0.11,2.40]$ and $[-2.40,-0.11]$) and the updated global fuzzy model were used. The RMSEs on $f(u(k))$ and $y(k)$ were even smaller than their counterparts in Fig.~\ref{fig:E41}, suggesting that adding a second patch fuzzy model was also beneficial. The PL models followed $f(u(k))$ and $y(k)$ immediately when the training started. Remarkably, it also followed the true model closely after $k=250$ when the training stopped, and after $k=500$ when the input dynamics changed dramatically.

Similar to Example~1, the performance of Bagging was identical to or worse than PL. LSBoost had smaller RMSE than PL when $L=0,1$, but PL outperformed LSBoost when $L=2$.

\subsection{Experiment 5: Mackey-Glass Chaotic Time Series Prediction}

Experiment~5 applied PL to Mackey-Glass chaotic time series prediction, similar to Example~4 in \cite{Jang1993}. The time series was generated by the chaotic Mackey-Glass differential delay equation \cite{Mackey1977}:
\begin{align}
\dot{x}(t)=\frac{0.2x(t-\tau)}{1+x^{10}(t-\tau)}-0.1x(t), \label{eq:MG}
\end{align}
discretized by the fourth-order Runge-Kutta method at $t=0,1,...,1117$. $x(0)=1.2$ and $\tau=17$ was used\footnote{According to \cite{Mackey1977}, (\ref{eq:MG}) is chaotic when $\tau>17$. However, \cite{Jang1993}, which is widely cited, used $\tau=17$, and we followed its practice. This dataset is also available in the Matlab Fuzzy Logic Toolbox under the name `mgdata.dat'.}. The goal was to use some known values of the time series up to $x(t)$ to predict its future value $x(t+P)$. More specifically, we used $(x(t-12),x(t-6),x(t))$ to predict $x(t+6)$. The first 617 points were used as the training data, and the last 500 as test. All global and patch fuzzy models in PL used ANFIS with two trapezoidal input MFs.

\begin{figure}[htbp]\centering
\subfigure[]{\label{fig:MG0}   \includegraphics[width=.92\linewidth,clip]{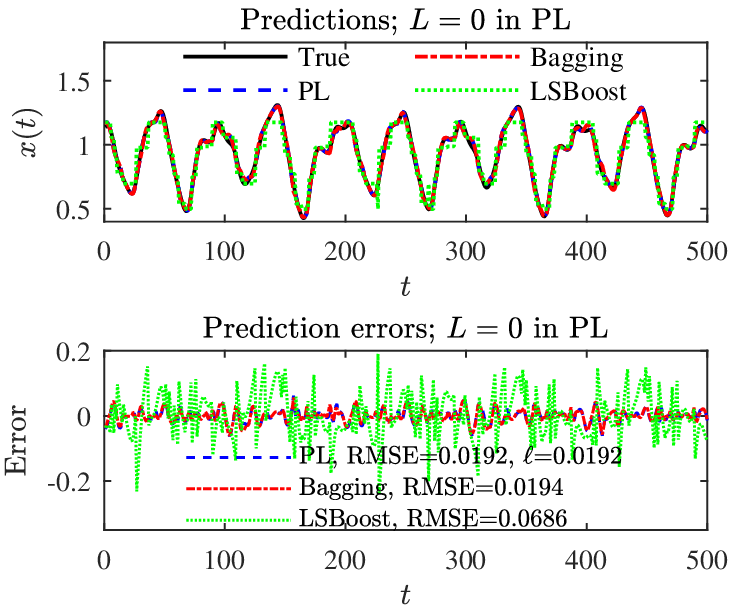}}
\subfigure[]{\label{fig:MG1}   \includegraphics[width=.92\linewidth,clip]{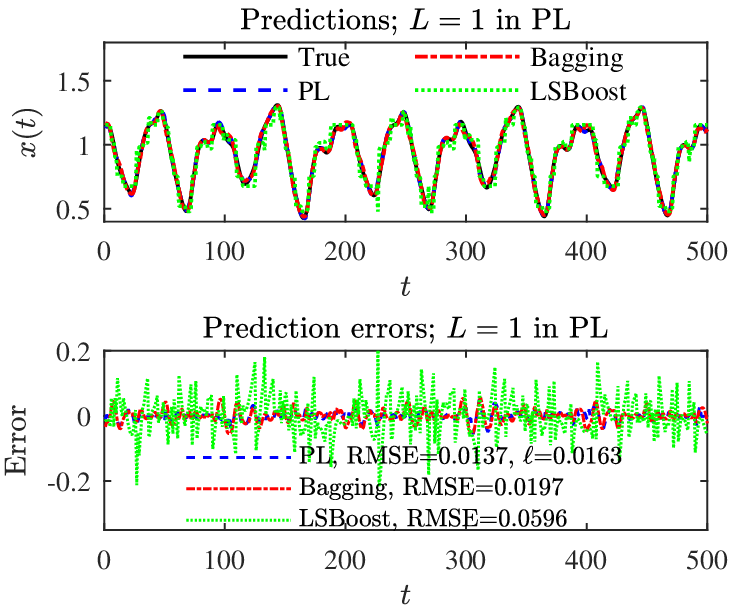}}
\subfigure[]{\label{fig:MG2}   \includegraphics[width=.92\linewidth,clip]{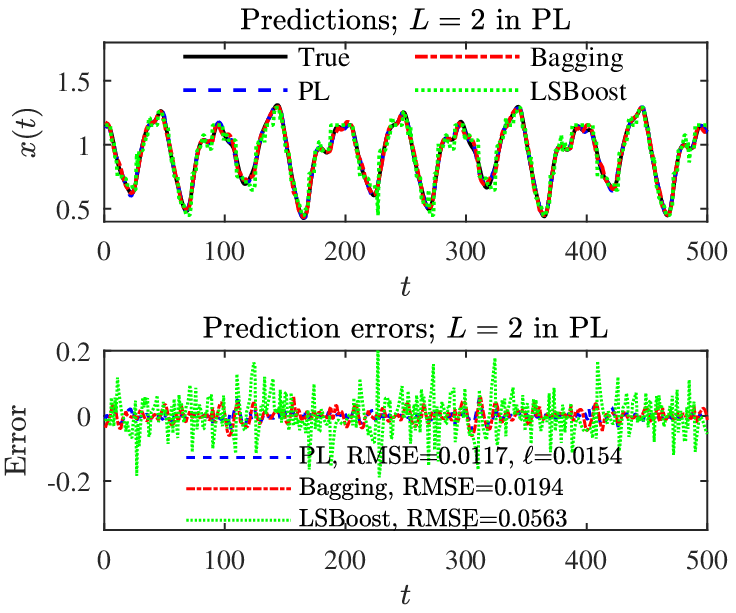}}
\caption{Experiment~5: Mackey-Glass chaotic time series prediction using PL, Bagging and LSBoost. (a) prediction results and errors using $L=0$ in PL, and one weak learner in Bagging and LSBoost; (b) prediction results and errors using $L=1$ in PL, and two weak learners in Bagging and LSBoost; and, (c) prediction results and errors using $L=2$ in PL, and three weak learners in Bagging and LSBoost.} \label{fig:MG}
\end{figure}

The test results are shown in Fig.~\ref{fig:MG}, for different $L$ in PL, and different numbers of weak learners in Bagging and LSBoost. The first three patches in PL were $[0.64,1.06]\times[0.57, 1.16]\times[0.99,2.08]$, $[0.64,1.06]\times[0.57, 1.16]\times[0.62,0.99]$, and $[1.06,2.08]\times[0.57, 1.16]\times[0.62,0.99]$, respectively. The RMSE and loss of PL decreased as $L$ increased from zero to two, as shown in the figure. However, as we further increased $L$ to three (not shown in the figure because it will be too long to be put in a single page), the RMSE decreased from 0.117 to 0.115, but the loss increased from 0.154 to 0.163. So, the optimal number of patches should be two in this problem.

It is also easy to observe from Fig.~\ref{fig:MG} that generally PL outperformed Bagging and LSBoost, especially when $L>0$ (the number of weak learners was larger than one).

\section{Limitations and Future Research} \label{sect:limitations}

Five experiments in the previous section demonstrated the feasibility of PL. However, we have to admit that PL is a brand-new concept, and significant future research is needed to make it more robust and powerful:
\begin{enumerate}
\item \emph{How to automatically identify the patches?} For a model with one or two inputs, we can visualize the errors and determine where patches should be put. However, visual examination is not feasible when there are more than two inputs. This paper focused on fuzzy systems using piecewise linear MFs, because the first-order rule partitions can be easily identified, and used as patch candidates by means of (\ref{eq:k})-(\ref{eq:kM}) (this is a huge advantage for a fuzzy system). It is desirable to have an automatic patch identification approach for other powerful machine learning models\footnote{Note that once a patch is identified, the corresponding patch model can be constructed by any machine learning algorithm, not necessarily a fuzzy system. So, the fact that currently we can only initialize the patches from first-order rule partitions of a fuzzy system does not limit the applications of PL.}, e.g., neural networks, support vector machines, etc. One possibility is to perform change point detection, e.g., using RuLSIF \cite{Liu2013}, on the training errors of the initial global model, and then identify the regions that give the largest errors.

\item \emph{Should the patches be generated adaptively?} In this paper we use the initial global model to generate a fixed patch pool, which may be viewed as a \emph{static} patch generation approach, i.e., once the patch pool is generated from the initial model, a patch may be removed from it, as explained in Algorithm~1, or stay in the patch pool as a candidate, but no patch candidate in the pool will be changed. However, it is also possible to generate the patches \emph{adaptively}, i.e., generate a global model, identify the patch pool, train the first patch model, update the global model, then identify a new patch pool according to the performance of the updated global model and the first patch model, and so on. The main difference between static PL and adaptive PL is whether the patch pool is updated when a new patch model is trained. Static PL is simpler because the patch pool is only constructed once, but adaptive PL may achieve better performance.

\item \emph{How to ensure the continuity at the patch boundaries?} At the patch boundary, a patch model may switch to another patch model, or to the global model. Generally this results in discontinuity at the patch boundaries. Discontinuities may not be a big issue for the five experiments in this paper, but may cause significant problems in other applications, e.g., controller design \cite{drwuCont2011}, in which discontinuities may result in system instability. So, it is favorable to design PL models that are always continuous. One possible solution is to use the training examples at the patch boundaries in training the corresponding patch models and also in updating the global model. Furthermore, larger weights may be assigned to these training examples to make sure different models give almost identical outputs at the boundary.

\item \emph{How to deal with high-dimensional data?} High-dimensional data are always challenging in machine learning, known as the curse of dimensionality. This problem is particularly significant in PL, because each patch model is trained using only the training data falling into the corresponding patch. As the dimensionality increases, the number of training data in a patch generally decreases, making the patch model training difficult. To alleviate this problem, one could start with simple patch models (e.g., linear regression, instead of nonlinear regression), or use regularization on complex models. However, more systematic and powerful approaches for dealing with high-dimensional data are needed.

\end{enumerate}
For complex problems, it may also be beneficial to perform further PL within a patch, and hence design a hierarchical (deeper-learned) PL model.

Additionally, when specific to PL using fuzzy systems, the following improvements can also be considered:
\begin{enumerate}
\item \emph{How to better determine the patches for trapezoidal or triangular MFs?} In this paper we use the first-order rule partitions as the patch candidates, and empirically showed that they work well. However, there is no guarantee that they are the best candidates. Additionally, instead of considering each partition individually, the combination of successive partitions may also be considered. For example, in Fig.~\ref{fig:Partition}, instead of considering only $P(k|x_m)$, $k=1,...,5$, as the five patch candidates, we may also consider $P(1|x_m)\cup P(2|x_m)$, $P(1|x_m)\cup P(2|x_m)\cup P(3|x_m)$, etc., as patch candidates.

\item \emph{How to determine the patches for Gaussian or bell MFs?} Gaussian and bell MFs have different first-order rule partition characteristics than trapezoidal and triangular MFs \cite{Mendel2018}. Because Gaussian and bell MFs span the entire universe of discourse, fuzzy systems that use them have only one first-order rule partition. Meanwhile, these MFs, particularly Gaussian MFs, have been widely used in practice. How to identify the patches for such MFs is an open yet very important question.

\item \emph{How to determine the optimal number and shape of MFs in the global and patch fuzzy models?} For simplicity, in each of our experiments the global and patch fuzzy models used the same number of trapezoidal MFs. However, they do not need to. Furthermore, the global and patch models do not need to use the same MF shape (e.g., one can use trapezoidal MFs, and another can use Gaussian MFs; one can use type-1 MFs, and another can use interval type-2 MFs). It is desirable to develop a systematic approach for determining the optimal number and shape of MFs in the global and patch fuzzy models\footnote{Of course, other algorithms such as the neural networks can also be used to construct the patch models. Here we assume that the user has determined to use fuzzy systems as the patch models.}. Cross-validation may be considered.

\item \emph{How to perform PL using interval or general type-2 fuzzy systems?} This paper focused on type-1 fuzzy systems, but numerous studies have demonstrated that interval and general type-2 fuzzy systems may be better able to cope with uncertainties than type-1 fuzzy systems \cite{Mendel2017}. It would be interesting to perform PL using interval or general type-2 fuzzy systems, as better performance is expected.
\end{enumerate}

Finally, this paper only considered regression problems. PL for classification problems are also very worthy of investigation.

\section{Conclusions} \label{sect:conclusion}

When the performance of a machine learning model is not satisfactory, there are different strategies to improve it, including designing a deeper, wider, and/or more nonlinear model, and ensemble learning (aggregate multiple base/weak learners in parallel or in series). This paper has proposed a novel strategy, \emph{patch learning}, to improve the performance of a machine learning model. It first identifies a few patches from the initial model, which contribute the most to the learning error. Then, a (local) patch model is trained for each such patch, using only training examples falling into the patch. Finally, a global model is trained using training data that do not fall into any patch. To use the PL model, we first determine if the input falls into any patch. If yes, then the corresponding patch model is used to compute its output. Otherwise, the global model is used. Five experiments on different regression problems verified the effectiveness of the proposed PL approach: as more patch models are added, the overall RMSE decreases. We also defined a loss function for determining the optimal number of patches, considering the trade-off between RMSE and model complexity.

It should be emphasized that although this paper focuses on PL using fuzzy systems, the idea is generic: any machine learning algorithm can be used as the patch model, and the patch models can also be trained using different machine learning algorithms.

%\bibliographystyle{IEEEtran}\bibliography{drwubib}

% Generated by IEEEtran.bst, version: 1.14 (2015/08/26)

\end{document}